%% file: root.tex
\documentclass[journal,twoside]{IEEEtran}
\usepackage{amsmath,amsfonts}
\usepackage{amssymb} %
\usepackage[caption=false,font=normalsize,labelfont=sf,textfont=sf]{subfig}
\usepackage{graphicx}
\usepackage{cite}
\usepackage{siunitx}
\usepackage{hyperref}
\usepackage{xcolor}
\hypersetup{
    colorlinks,
    linkcolor={red!50!black},
    citecolor={blue!50!black},
    urlcolor={blue!80!black}
}

\usepackage{pgfplots}
\graphicspath{{./img/}}

\everymath=\expandafter{\the\everymath\displaystyle}

\usepackage{mathtools}
\usepackage{booktabs}
\usepackage{makecell}

\let\vec\boldsymbol

\newcommand{\mlpgap}{\mathrm{MLP}_\mathrm{gap}}
\newcommand{\mlpacc}{\mathrm{MLP}_\mathrm{acc}}
\newcommand{\mlpurg}{\mathrm{MLP}_\mathrm{urg}}

\newcommand{\veh}{\mathrm{V}}

\newcommand{\prioset}{\mathcal{P}}
\newcommand{\vehcons}{c}
\newcommand{\swcompl}{\mathrm{c}}
\newcommand{\effcost}{\mathrm{L}}
\newcommand{\urg}{\mathrm{u}}

\usepackage[absolute,overlay,showboxes]{textpos}
\TPMargin{1mm}
\newcommand\copyrighttext{
	\footnotesize
	\noindent
	SUBMITTED TO REVIEW AND POSSIBLE PUBLICATION. COPYRIGHT WILL BE TRANSFERRED WITHOUT NOTICE.\\
	Personal use of this material is permitted.
	Permission must be obtained for all other uses, in any current or future media, including reprinting/republishing this material for advertising or promotional purposes, creating new collective works, for resale or redistribution to servers or lists, or reuse of any copyrighted component of this work in other works.}%
\newcommand\copyrightnotice{%
	\textblockcolour{white}%
	\begin{textblock*}{7.12in}(0.68in,0.15in)
		\copyrighttext
	\end{textblock*}
}

\begin{document}

\title{Generalized Coordination of \\ Partially Cooperative Urban Traffic}

\author{Max Bastian Mertens and Michael Buchholz%
	\thanks{This research has been conducted as part of the PoDIUM project, which is funded by the European Union under grant agreement No. 101069547. Views and opinions expressed are however those of the authors only and do not necessarily reflect those of the European Union or European Commission. Neither the European Union nor the granting authority can be held responsible for them.}%
	\thanks{The authors are with the Institute of Measurement, Control and Microtechnology,
		Ulm University, D-89081 Ulm, Germany.
		E-mail addresses: {\tt firstname.lastname@uni-ulm.de}
	}%
}

\markboth{IEEE Transactions on Intelligent Vehicles, Vol.~X, No.~Y, Month~2024}{Mertens \MakeLowercase{\textit{et~al.}}: Generalized Coordination of Partially Cooperative Urban Traffic}

\maketitle
\copyrightnotice

\begin{abstract}
Vehicle-to-anything connectivity, especially for autonomous vehicles, promises to increase passenger comfort and safety of road traffic, for example, by sharing perception and driving intention.
Cooperative maneuver planning uses connectivity to enhance traffic efficiency, which has, so far, been mainly considered for automated intersection management.
In this article, we present a novel cooperative maneuver planning approach that is generalized to various situations found in urban traffic.
Our framework handles challenging mixed traffic, that is, traffic comprising both cooperative connected vehicles and other vehicles at any distribution.
Our solution is based on an optimization approach accompanied by an efficient heuristic method for high-load scenarios.
We extensively evaluate the proposed planer in a distinctly realistic simulation framework and show significant efficiency gains already at a cooperation rate of 40\,\%.
Traffic throughput increases, while the average waiting time and the number of stopped vehicles are reduced, without impacting traffic safety.
\end{abstract}


\section{Introduction}

Intelligent infrastructure has long been used to actively manage traffic using adaptive traffic light switching.
As urban traffic recently showed an increasing amount of vehicle-to-vehicle and vehicle-to-infrastructure communication, connected vehicles can also contribute to a collective environment model.
This passively increases traffic safety and efficiency by providing improved perception supported by edge computing~\cite{buchholz_handling_2021}.
With more and more connected automated vehicles (CAVs) prevalent, individual vehicles can be actively guided for more fine-grained control than with static rules or traffic lights:
When two or more cooperative CAVs approach a crossing or merging scenario from different lanes, a cooperative maneuver can resolve the conflict in the most favorable way by temporarily deviating from static right-of-way rules or traffic light phases.
Such an automated intersection management (AIM) can improve traffic efficiency at single unsignalized intersections, as shown in our prior work~\cite{mertens_cooperative_2022} and many others.

Most of the research on AIM assumes fully connected automated traffic to reduce uncertainties and algorithm complexity.
Fewer works focus on mixed traffic, that is, cooperative CAVs and non-cooperative human-driven vehicles (HDVs) together on the same roads.
The handling of mixed traffic is much more difficult, as the future behavior of HDVs, such as gap acceptance and turning direction, is unknown and cannot be controlled.
To retain safety, all or the most probable actions of HDVs have to be considered, drastically increasing the planning complexity and reducing the number of feasible cooperative maneuvers.
In addition, handling mixed traffic requires a realistic driver model, that is, a prediction of HDV behavior in the foreseeable future during the whole maneuver planning horizon, which usually extends over at least 10\,s.

\begin{figure}
	\centering%
	{
	\def\svgwidth{0.995\linewidth}
	\small
	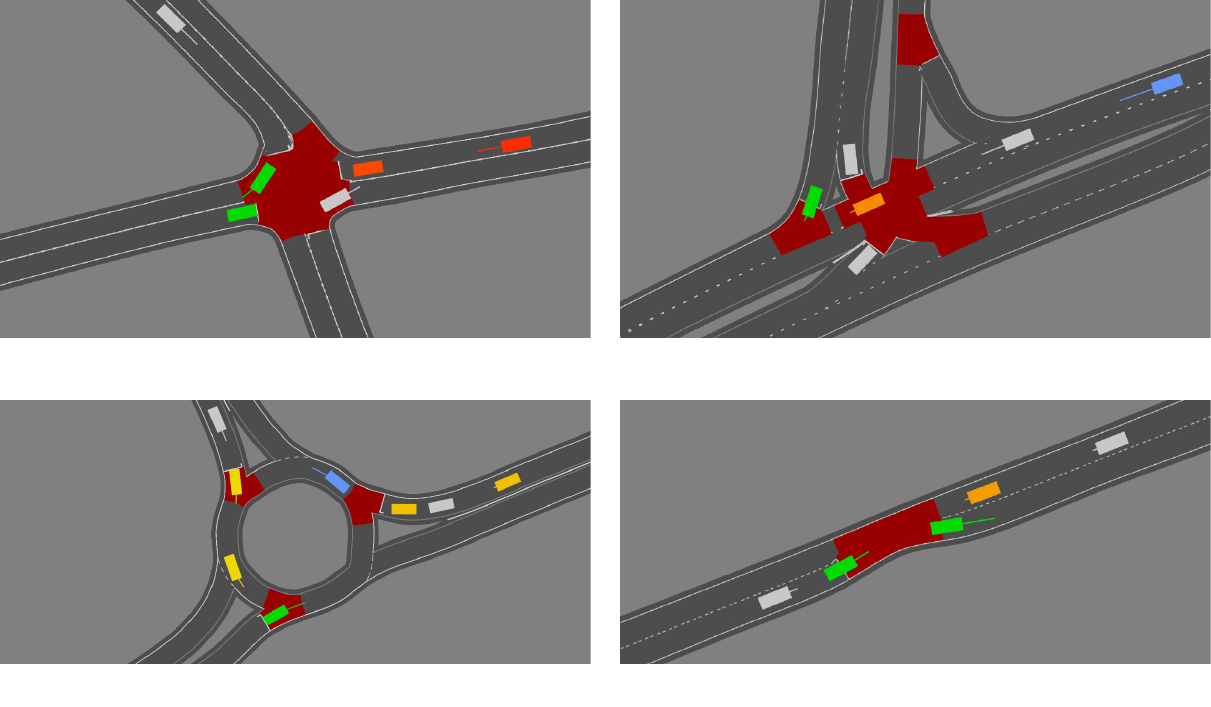\vspace{-0.1cm}~
	}
	\caption{
        Four of the 13 simulated real-world scenarios. They comprise five intersections with right-before-left regulation~(a) or main and side roads~(b), seven roundabouts~(c), and a road narrowing~(d).
        Green/yellow/red vehicles are coordinated CAVs with early/medium/late entrance time;
        uncoordinated CAVs are blue; gray vehicles are HDVs.
        Dark red areas denote conflict zones.
	}
	\label{fig:intro_scenarios}
\end{figure}

In our work, we explicitly tackle mixed traffic with a large proportion of HDVs as prevalent in real-world traffic.
For this, we designed a vehicle behavior and motion prediction and devised a centralized cooperative maneuver planning for unsignalized intersections~\cite{mertens_fast_2024}.
We achieved increased traffic throughput and reduced waiting times even with low CAV percentages, demonstrated by extensive simulations and test drives in public traffic~\cite{mertens_cooperative_2022,klimke_real-world_2025}.
In this article, we generalize our prediction and cooperative maneuver planning approach to support a whole class of merging and crossing situations found in urban mixed traffic, shown in Fig.~\ref{fig:intro_scenarios} and subsequently called \emph{intersection-like scenarios}.
We complement our optimization maneuver search with a faster, but suboptimal, planning heuristic satisfying real-time requirements even in high-load scenarios. %
We demonstrate the effectiveness of both planning methods through extensive simulations of different unsignalized intersection types, roundabouts, and a road narrowing.
Our novel mixed traffic simulation framework combines a state-of-the-art prediction model for HDVs~\cite{strohbeck_deepsil_2021}, real-world motion planning and control modules for CAVs~\cite{ruof_real-time_2023}, and a real-world maneuver coordination protocol for V2X communication~\cite{mertens_extended_2021}.
Our scientific contribution thus comprises:
\begin{itemize}
\item Enhancing our driver model to excel on various datasets,
\item Generalizing our cooperative maneuver planning method to support various urban intersection-like scenarios,
\item Introducing a performant heuristic planning method, and
\item Quantitatively evaluating both methods in a novel realistic mixed traffic simulation framework.
\end{itemize}

The remainder of the paper is structured as follows:
Section~\ref{sec:rel_work} discusses related works in cooperative planning for automated and mixed traffic in intersection-like scenarios.
Our approach is detailed in Section~\ref{sec:approach}.
We describe our experiments and simulation framework and discuss our evaluation results in Section~\ref{sec:experiments}.
Finally, Section~\ref{sec:conclusion} summarizes the article and gives an outlook on future work.

\section{Related Work}
\label{sec:rel_work}

A lot of research has already been published on traffic control at intersections.
One of the first frequently cited papers on AIM~\cite{dresner_multiagent_2004} was published 20 years ago.
Since then, many research groups have investigated cooperative maneuver planning methods at various intersection types.

Early works on AIM often focus on simpler scenarios, such as fully cooperative traffic, and evaluate in simplistic simulation environments, as surveyed by~\cite{zhong_autonomous_2021}.
When all vehicles can be controlled, the most efficient crossing order at an intersection can be found by enumeration~\cite{li_cooperative_2006}.
However, the number of combinations increases exponentially with the number of vehicles~\cite{wu_cooperative_2012}, so this approach is not computationally feasible in a real-time system.
Thus, algorithms approximating the optimal solution have been developed.
Different researchers proposed a first-in, first-out (FIFO) heuristic to directly generate a suboptimal crossing order~\cite{dresner_multiagent_2004,wuthishuwong_vehicle_2013}.
This already significantly increases traffic efficiency in a 100\,\%~CAVs environment.
More elaborate approaches find a near-optimal vehicle crossing order using optimization algorithms and data structures, such as ant colonies~\cite{wu_cooperative_2012}, Petri nets~\cite{ahmane_modeling_2013}, or Monte Carlo Tree Search~\cite{kurzer_decentralized_2018}.
All works mentioned above are only capable of handling fully automated traffic.
Most approaches are only tested in a simplistic simulation without considering processing delays; only the authors of~\cite{ahmane_modeling_2013} provide a run time analysis and real-world experiments.

Fewer works have been published about the traffic management at other intersection-like scenarios such as roundabouts.
In~\cite{bichiou_developing_2019}, a roundabout management approach based on optimal control theory is proposed.
The large computational cost of calculating the optimal solution makes the algorithm impractical for real-world usage.
Several real-time capable heuristics have been investigated in~\cite{Martin-Gasulla_traffic_2021} and a large increase in traffic throughput is reported.
An early work has been presented in~\cite{bento_intelligent_2012}, providing a coordinated FIFO ordering of vehicles through an intersection next to a roundabout.
However, all of those approaches depend on fully automated and connected traffic.

Only some researchers have considered AIM algorithms for mixed traffic at intersections or roundabouts.
Previous approaches for fully automated traffic were complemented by traffic light controls to support HDVs~\cite{dresner_sharing_2007,bento_intelligent_2013}.
However, these methods only work well for a low percentage of HDVs ($\le$\,10\,\%).
Other works additionally use CAVs as virtual platoon leaders to indirectly control larger amounts of HDVs~\cite{qian_priority-based_2014,yang_isolated_2016}, but still rely on the presence of traffic lights.
In~\cite{zhao_optimal_2018}, a FIFO heuristic is investigated for mixed traffic at roundabouts.
However, in this work, the authors also report insubstantial benefits of improved vehicle ordering for less than 100\,\%~CAVs.
Furthermore, many of the aforementioned approaches assume knowledge about the route, i.e. the turn direction of HDVs, which is generally not given in public traffic.

Some works are close to our problem formulation and approach.
The research published in~\cite{nichting_space_2020} proposes a coordination procedure for conflicts at generalized shared road segments in various intersection-like scenarios using a dedicated V2X protocol.
The method works in mixed traffic, but coordinates between only two CAVs in a decentralized way, resulting in a problem scope different from our research.
A recent publication~\cite{klimke_automatic_2023} proposes a reinforcement learning solution with graph neural networks to handle mixed traffic of any HDV rate at intersections while handling unknown HDV routes.
The method is shown to being real-world capable and to significantly reduce waiting times for CAVs and HDVs.
However, the approach is limited in generalizing to intersection layouts that have not been encountered during training.

Our previously published AIM approach~\cite{mertens_cooperative_2022} is specifically designed to support any HDV rate and unknown HDV routes.
There, a real-time capable optimization algorithm improves the crossing order of conflicting CAV pairs to lower the expected overall delay.
A complementing scene-consistent multi-scenario prediction has been published in~\cite{mertens_fast_2024} to provide a realistic estimate of the expected efficiency gain.
Consequently, in this work, we combine the two methods and extend both to support other intersection-like scenarios such as roundabouts and road narrowings.
We evaluate in a realistic simulation framework and compare our optimization approach with different heuristics.
To our knowledge, there is no comparable scientific publication on generalized central management of mixed traffic at various types of scenarios with a realistic and detailed quantitative evaluation.

\section{Approach}
\label{sec:approach}

This section first gives an overview of the system considered to derive the problem formulation.
Subsequently, the proposed method and the details of each part of the algorithm are described.

\subsection{Problem Overview}

Our problem scope is a single traffic scene in an urban road network, such as an intersection, a roundabout, or a road narrowing, with one or more \emph{conflict zones}.
These zones are road segments shared by vehicles coming from different directions (cf.~Fig.~\ref{fig:intro_scenarios}), e.g., at merging or crossing lanes.
The beginning of such a zone is called the stop line, while the exit is called target line.
Zone geometries are extracted from map data so that collisions are prevented if only vehicles from one direction occupy a conflict zone at any time.

\subsubsection{System Overview}
\begin{figure}
	\centering%
	{
	\def\svgwidth{0.8\linewidth}
	\small
	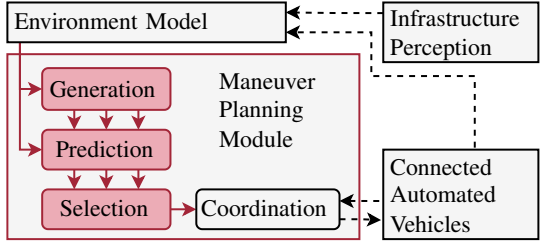
	}
	\caption{
        System overview. Parts handled in this work are marked in red, dashed lines denote V2X communication.
        The two modules on the left are deployed on an edge server close to the road users and infrastructure on the right.
	}
	\label{fig:system_overview}
\end{figure}

An overview of the proposed maneuver planning framework is depicted in Fig.~\ref{fig:system_overview}.
We assume availability of a road-side infrastructure perception system that observes an approximately 100\,m diameter around the scene, such as in~\cite{buchholz_handling_2021}.
The perception hardware can be installed permanently or set up temporarily, such as during the presence of a construction site blocking parts of a lane.
The detections are fused into an environment model comprising the physical state of all road users, which is generated on an edge server near the scene.
Our scope of the problem focuses on only HDVs and CAVs in the scene.
Other road users, such as cyclists, could be considered employing a respective prediction model.
The CAVs present in the area are expected to provide their state and planning information to provide accurate knowledge of their location and route in the environment model.
A centralized cooperative maneuver planning module on the edge server optimizes the crossing and merging order between CAVs to improve the overall traffic flow.

The cooperative maneuvers are coordinated by a feedback loop using the V2X protocol proposed in~\cite{mertens_extended_2021}.
A maneuver is communicated to a vehicle $\veh_i$ as a set $\vehcons_i$ of longitudinal space-time constraints, similar to the approach in~\cite{nichting_space_2020}:
\begin{equation}
    \label{eq:s_t_cons}
    \vehcons_i = \left\{\left(s_1,t_\mathrm{min}(s_1),t_\mathrm{max}(s_1)\right),~\ldots\right\}.
\end{equation}
Each waypoint $s$ along the vehicle route needs to be crossed between $t_\mathrm{min}(s_1)$ and $t_\mathrm{max}(s_1)$.
Typically, vehicles are assigned a $t_\mathrm{min}$ constraint at the entrance (stop line) and a $t_\mathrm{max}$ constraint at the exit (target line) of a conflict zone.
As long as the time intervals of vehicles from different directions do not overlap, conflict zones are guaranteed to be collision-free.
CAVs are required to have a motion planning compatible with these space-time constraints, such as that presented in~\cite{ruof_real-time_2023}.
The CAV motion planning remains responsible for generating a trajectory and checking the viability of the requested maneuver.
Whenever a request is rejected, the maneuver is aborted for all vehicles and the planning is restarted.

\subsubsection{Route Knowledge}
\label{ssec:route_knowl}

As the route information is not directly shared between vehicles, CAVs would have to conservatively assume the most conflicting route of other vehicles and sometimes unnecessarily yield.
Therefore, for each CAV, the maneuver planning also sends a set of non-conflicting vehicles, i.e., all other CAVs that will not merge or cross.

\subsubsection{Problem Formulation}

The maneuver planning should be a software module on the edge server that processes the environment model and continuously generates maneuvers, that is, constraint sets $\vehcons_i$ for each CAV $\veh_i$ in the scene.
The objective is to reduce average waiting times, reduce stops, and increase traffic throughput at the observed scene.

As a fundamental constraint, maneuver planning should handle mixed traffic, that is, it should consider HDVs during the safety and efficiency analysis of maneuvers to not endanger or disadvantage HDVs.
In addition, it should handle various intersection-like scenarios.
For simulation and evaluation, we selected and mapped 13 diverse real-world traffic scenarios.
They comprise five intersections with main and side roads, one intersection with right-before-left regulation, seven roundabouts, and a road narrowing with priority from one direction.
For four intersections and all roundabouts, real-world traffic data from drone observations is available~\cite{bock_ind_2020,breuer_opendd_2020}.

\subsection{Approach Overview}

Our proposed maneuver planning framework is based on our previous method introduced in~\cite{mertens_cooperative_2022}.
In this article, we integrate our scene-consistent prediction module presented in~\cite{mertens_fast_2024} and generalize the method to different intersection-like scenarios.

The maneuver planning module operates cyclically at a frequency of 5\,Hz.
As a pre-processing step, all vehicles in the environment model are associated and projected to the centerlines of an internal map format.
The routes, i.e., the turning directions are known for CAVs but not for HDVs.
Thus, as a conservative assumption, the most conflicting possible route is expected for each HDV.
Likewise, HDVs do not know the turning directions of other vehicles, so the prediction of HDVs assumes a conflict with any crossing or merging lanes as well.

A maneuver is internally represented as a set $\prioset$ of priority pairs.
Each element is a pair of CAVs $\langle\veh_i,\veh_j\rangle$ where the first is prioritized over the second.
In each cycle at discrete time $k$, it performs the following steps, as already indicated in Fig.~\ref{fig:system_overview}:
\begin{enumerate}
    \item generation of new priority sets $\prioset_{k,1},\ldots,\prioset_{k,n}$,
    \item multiple predictions of the traffic scene when obeying the previously best $\prioset_{k-1}^\star$ or one of the new priority sets,
    \item selection of the new most favorable priority set $\prioset_{k}^\star$ and extraction of vehicle constraints $\vehcons_{i,k}$.
\end{enumerate}
The underlying time loss metric $\effcost(\prioset_k)$ to evaluate potential maneuvers $\mathcal{P}_k$ is defined as the relative velocity loss integrated over the predicted time horizon summed over all vehicles \cite{mertens_cooperative_2022}:
\begin{align}
	\label{eq:eff_simple}
	\effcost(\mathcal{P}_k) &\coloneqq \sum_{\veh_i} w_i\int_{T_\mathrm{start}}^{T_\mathrm{end}} 1-\frac{v_i^{(\prioset_k)}(t)}{v_{\max,i}^{(\prioset_k)}(t)} \,\mathrm dt,
\end{align}
where $v_{\max,i}(t)$ denotes the speed limit at the position of $\veh_i$ at time $t$ and $w_i$ is a weighting factor.
The method steps are detailed in the following sections.

\subsection{Priority Set Generation}

We devised an optimization priority set generation for medium load scenarios and a faster suboptimal heuristic method for real-time performance even in high traffic load.
For comparison, we provide two simple baselines and the non-cooperative case (called \emph{none}).

\subsubsection{Optimization Priority Set Generation}

The optimization method (OPT) generates numerous priority sets based on the previously best solution $\prioset_{k-1}^\star$.
The following options are gathered as potential new priority sets:
\begin{itemize}
    \item aborting the current maneuver, i.e., $\prioset_{k}=\emptyset$,
    \item removing single priority pairs: $\prioset_{k}=\prioset_{k-1}^\star\backslash\left\{\langle\veh_i,\veh_j\rangle\right\}$ for each $\langle\veh_i,\veh_j\rangle\in\prioset_{k-1}^\star$,
    \item reversing single priority pairs: $\prioset_{k}=\left(\prioset_{k-1}^\star\backslash\left\{\langle\veh_i,\veh_j\rangle\right\}\right)\linebreak\cup\left\{\langle\veh_j,\veh_i\rangle\right\}$ for each $\langle\veh_i,\veh_j\rangle\in\prioset_{k-1}^\star$,
    \item adding single priority pairs: $\prioset_{k}=\prioset_{k-1}^\star\cup\left\{\langle\veh_i,\veh_j\rangle\right\}$ or $\prioset_{k}=\prioset_{k-1}^\star\cup\left\{\langle\veh_j,\veh_i\rangle\right\}$ for each conflicting CAV pair $\langle\veh_i,\veh_j\rangle\notin\prioset_{k-1}^\star$,
    \item adding two or more priority pairs (equations omitted).
\end{itemize}
The total number of predicted scenarios is limited to 100, keeping the planner runtime lower than the target cycle time of 200\,ms in more than 97\,\% of the runs.
When the cycle time is exceeded, the previous maneuver is pursued for another cycle.

\subsubsection{Heuristic Priority Set Generation}

For high-load scenarios, we provide a suboptimal heuristic approach (HEUR).
We trained a small multi-layer perceptron (MLP) with two hidden layers of size 16 and $\tanh$ activation function to estimate an urgency metric $\urg_i$ for a CAV $\veh_i$ depending on its current state:
\begin{equation}
    \urg_i = \mlpurg(d_{\mathrm{stop},i}, v_i, n_{\mathrm{lead},i}, n_{\mathrm{foll},i}),
\end{equation}
where $d_{\mathrm{stop},i}$ is the longitudinal distance to the beginning of the conflict zone, $v_i$ is the velocity, $n_{\mathrm{lead},i}$ is the number of lead vehicles, and $n_{\mathrm{foll},i}$ is the number of following vehicles.
During maneuver planning, the conflicting CAV pairs are then ordered by urgency by defining respective priority pairs:
\begin{equation}
    \prioset_{k}=\left\{\langle\veh_i,\veh_j\rangle \,\middle|\, \urg_i > \urg_j \right\}.
\end{equation}

The MLP is trained on a dataset generated during 2000 simulation runs with a length of 120\,s each with 10 CAVs randomly spawned in a random scenario type.
In each time step $k$, a random conflicting CAV pair is chosen and both $\prioset_{k,ij}=\left\{\langle\veh_i,\veh_j\rangle\right\}$ and $\prioset_{k,ji}=\left\{\langle\veh_j,\veh_i\rangle\right\}$ are evaluated on a prediction horizon of 10\,s.
Each such evaluation generates a training label based on the negated cost metric from Eq.~\eqref{eq:eff_simple}:
\begin{equation}
    \urg_i - \urg_j = -\effcost(\mathcal{P}_{k,ij}) + \effcost(\mathcal{P}_{k,ji}).
\end{equation}
In this way, the MLP learns an abstract representation of the influence of a single vehicle on the efficiency of the total scene.

\subsubsection{Baseline: FIFO Priority Set Generation}

As a baseline, we order the CAVs arriving at the scene in a FIFO scheme.
The method is similar to the HEUR approach, but the urgency metric is defined as a simple prediction of the time until reaching the next conflict zone:
\begin{equation}
    \label{eq:fifo_urgency}
    \urg_i = \frac{d_{\mathrm{stop},i}}{\max\{v_i, \SI{0.1}{ms^{-1}}\}}.
\end{equation}
This baseline is comparable to the approach in~\cite{nichting_space_2020}.

\subsubsection{Baseline: Non-Conflicting Vehicles}

To distinguish the efficiency gains provided by route knowledge (cf. Section \ref{ssec:route_knowl}), we provide another NC (non-conflicting) baseline where no maneuvers are planned, but still the list of non-conflicting CAVs is communicated.

\subsection{Scenario Prediction}

To evaluate and compare potential new priority sets, the traffic scene is predicted on a 12\,s horizon for each $\prioset_{k}$, while the considered priority set is executed by the predicted CAVs.
For the first 1\,s, the previously selected $\prioset_{k-1}^\star$ is still followed, to account for communication and processing delays.

\subsubsection{Prediction Model}

To predict the behavior of HDVs and CAVs in the scene, we employ an extended version of the approach in~\cite{mertens_fast_2024}.
The prediction model is based on a neural network architecture comprising two MLPs, $\mlpacc$ and $\mlpgap$, to model the gap acceptance and acceleration, respectively.
The MLPs have two hidden layers of size 16 and a $\tanh$ activation function and were trained on real-world traffic data.
The input and output features of both stages are listed in Table~\ref{tbl:agent_observation}.

In each prediction time step, $\mlpgap$ estimates the gap acceptance, i.e., whether a low-priority vehicle will enter a conflict zone if prioritized vehicles approach the same area.
The gap acceptance estimation compares the velocity $v_i$ and the target line distance $d_{\text{targ},j}$ of a low priority vehicle $\veh_i$ to the velocity $v_j$ and stop line distance $d_{\text{stop},j}$ of a prioritized conflicting vehicle $\veh_j$ as described in~\cite{mertens_fast_2024}.
The gap acceptance can be overridden by the pairs in the considered priority set.

\begin{table}[t]
	\vspace{0.2cm}
	\caption{Observation input features of the two prediction MLPs.\linebreak
		Table adapted from~\cite{mertens_fast_2024}.}
	\label{tbl:agent_observation}
	\begin{center}
		\begin{tabular}{l l}
			\toprule
			\thead[l]{Environment observation $\vec{o}_{i}$ of $\veh_i$} & \thead[l]{Input features} \\
			\midrule
			Distance to stop line & $d_{\text{stop},i}$ \\
			Current and maximum velocity & $v_i$, $v_{\max,i}$ \\
			Max. upcoming abs. lane heading diff. & $\Delta\psi_{\max,i}$ \\
			Lead vehicle distance and velocity & $d_{\text{lead},i}$, $v_{\text{lead},i}$ \\
			\toprule
			\thead[l]{Gap obs. $\vec{o}_{i,j}$ of $\veh_i$ towards $\veh_j$} & \thead[l]{Input features} \\
			\midrule
			Distance to target line & $d_{\text{targ},i}$ \\
			Velocity & $v_i$ \\
			Distance of other vehicle to stop line & $d_{\text{stop},j}$ \\
			Velocity of other vehicle & $v_j$ \\
			\bottomrule
		\end{tabular}
	\end{center}
\end{table}

The driver model $\mlpacc$ uses the gap acceptance as well as state and environment observations to estimate the acceleration.
The acceleration of each vehicle is integrated to generate the vehicle states of the next time step.
The inputs to $\mlpacc$ were modified compared to~\cite{mertens_fast_2024}:
The multiple previous and upcoming relative lane headings were replaced by a maximum absolute difference between the current and upcoming headings during the next 100\,m.
In addition, two gap acceptance related features were added: the velocity $v_j$ and the stop line distance $d_{\text{stop},j}$ of the closest conflicting vehicle $\veh_j$.
In this way, a vehicle at a soon-to-be cleared conflict zone can already accelerate while the gap acceptance still reports a required stop.

\subsubsection{Model Training and Evaluation}

We adopted the training schemes from~\cite{mertens_fast_2024} and extended the training data to all four intersections in the inD dataset~\cite{bock_ind_2020} and added traffic data from five roundabouts from the openDD dataset~\cite{breuer_opendd_2020}.
For each of the nine scenarios, more than an hour of relevant traffic observations was extracted from the datasets.

The driver model $\mlpacc$ is pre-trained using Proximal Policy Optimization~\cite{schulman_proximal_2017} in a multi-agent closed-loop reinforcement learning environment on each of the 13 considered scenarios.
The manually designed objective function rewards driving close to the speed limit and avoiding collisions with prioritized vehicles, thus aiming at natural driving behavior.
The model is re-trained using Generative Adversarial Imitation Learning~\cite{ho_generative_2016} for more realistic behavior.

The gap acceptance decision model $\mlpgap$ is trained using supervised learning on precedence labels extracted from both datasets.
The data was resampled to use the same number of labels for intersections and roundabouts.

\begin{figure}[tb]
	\vspace{0.08cm}
	\centering
		\input{img/s_over_t.pgf}
	\caption{Prediction accuracy per vehicle of our $\mlpacc$ vs. IDM on the inD and openDD datasets.
		Our gap acceptance model $\mlpgap$ was used for both.
    }
	\label{fig:pred_eval}
\end{figure}
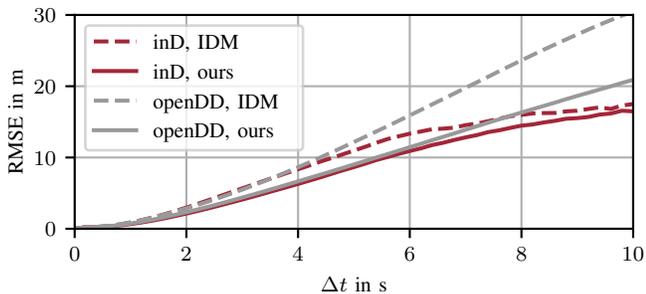

We evaluated the joint prediction consisting of both models.
We sampled one scenario per second in the datasets, predicted the scene over a horizon of 10\,s and compared the vehicle states with the ground truth.
On the inD dataset, our models predicted the correct vehicle crossing order in 84.4\,\% of the scenes, compared to 77.8\,\% on the openDD dataset.
In Fig.~\ref{fig:pred_eval}, we compare the prediction accuracy of our driver model $\mlpacc$ with the Intelligent Driver Model (IDM) while using our gap acceptance model $\mlpgap$ in both evaluations.
We use the RMSE of the predicted vehicle position compared to the ground truth of the respective dataset as an accuracy metric.
On the inD dataset, we achieve a position RMSE of 16.5\,m compared to the IDM result of 17.5\,m, both at the end of the prediction horizon of 10\,s.
On the openDD dataset, our method reaches an RMSE of 20.9\,m, while the IDM error is much higher at 30.5\,m.
The RMSE values are slightly higher than our reported value of 14\,m in~\cite{mertens_fast_2024}, which comes from the fact that we now generalize to nine different traffic scenarios instead of just one.

\subsection{Priority Set Selection and Maneuver Extraction}

Of the predicted priority sets, the most favorable one needs to be selected.
First, predictions with collisions due to infeasible priority pairs or those with unfulfilled priority pairs are discarded.
From the remaining predictions, the priority set with the most favorable maneuver metric is selected as the result $\prioset_k^\star$ of the current planning cycle.

\subsubsection{Maneuver Metric}
We use the time loss metric in Eq.~\eqref{eq:eff_simple} to evaluate potential maneuvers.
To reduce switching between different maneuvers in each cycle, the time loss is extended by a switching complexity term $\swcompl$ similar to~\cite{mertens_cooperative_2022}:
\begin{align}
	\label{eq:eff_full}
	\tilde{\effcost}(\mathcal{P}_k) &\coloneqq \effcost(\mathcal{P}_k) + \swcompl(\mathcal{P}_k, \mathcal{P}_{k-1}^\star) \\
    \text{with } \swcompl(\mathcal{P}_k, \mathcal{P}_{k-1}^\star) &\coloneqq 
        \SI{1}{s} \cdot \left|\mathcal{O}(\mathcal{P}_k)\backslash\mathcal{O}(\mathcal{P}_{k-1}^\star)\right|,
\end{align}
where $\mathcal{O}(\mathcal{P})$ is the predicted vehicle crossing order resulting from the imposed priorities $\mathcal{P}$, described as the set of ordered conflicting vehicle pairs.
Thus, a switched vehicle pair order must in turn lead to a time loss reduction of at least 1\,s.

To prevent the maneuver planning from stopping some vehicles forever to increase the overall traffic efficiency, a simple weighting of vehicles based on their waiting time is defined:
\begin{equation}
    \label{eq:time_loss_weight}
    w_i \coloneqq 1+\frac{t_{\mathrm{slow},i}}{\SI{10}{s}},
\end{equation}
where $t_{\mathrm{slow},i}$ is the time span since the velocity of vehicle $\veh_i$ has been less than 10\,km/h, or zero if $\veh_i$ is faster.

\subsubsection{Derivation of Cooperative Maneuvers}

As the chosen priority set is just an internal high-level description, it must be converted into space-time constraints $\vehcons_i$ for each CAV as in Eq.~\eqref{eq:s_t_cons}.
Therefore, for each pair of conflicting CAVs, the time of the prioritized vehicle leaving the conflict zone is extracted from the prediction.
This timestamp is then communicated as $t_\mathrm{max}$ of the prioritized CAV to cross the target line and as $t_\mathrm{min}$ of the yielding CAV to cross the stop line.
In this way, conflicting vehicles are ensured not to drive in the conflict zone at the same time, while leaving as much degree of freedom to the local trajectory planning as possible.

\section{Evaluation}
\label{sec:experiments}

To demonstrate effectiveness and real-world usability, we performed extensive simulations employing our maneuver planning approach in various scenarios and configurations.
A demonstrative video showing some of the simulated scenarios is available at \url{https://youtu.be/MSICDX76qaU}.

\subsection{Scenarios}

We sampled a total of 316 scenarios, distributed across four intersections from the inD dataset~\cite{bock_ind_2020}, seven roundabouts from the openDD dataset\cite{breuer_opendd_2020}, and one additional intersection as well as one road narrowing located in Ulm-Lehr, Germany.
The maps used for simulation were modeled from aerial images and contain road centerlines with speed limits and right-of-way regulations.
In each scenario, the number, routes, and spawn positions of the vehicles are randomly sampled so that the initial states do not cause collisions or violation of the right of way.
The number of HDVs and CAVs according to the target CAV percentage is randomly distributed among vehicles.

Each scenario is simulated for 60\,s for all maneuver planning methods and baselines (none, NC, FIFO, HEUR, OPT) and 11 different percentages of CAVs in the scene, producing a total of approximately 290 hours of traffic simulation.
As our CAV software stack is in a prototype state, there are some remaining unsolved edge cases.
Approximately 1.1\,\% of the simulation runs were not successful and resulted in a software crash or timeout, mostly unrelated to maneuver planning.
Those scenarios were excluded from all evaluations.

\subsection{Simulation Framework}

We use the DeepSIL simulation framework published in~\cite{strohbeck_deepsil_2021}.
It is based on the ROS2 software distribution and can simulate HDVs and CAVs jointly in a traffic scene.
We extended the framework to support multiple CAVs running the trajectory planning stack from~\cite{ruof_real-time_2023}.
Our cooperative maneuver planning approach is developed and integrated as a ROS2 software module.
The communication between the central planner and each simulated CAV is based on the V2X communication protocols proposed in \cite{buchholz_handling_2021,mertens_extended_2021}.

\subsubsection{Vehicle Models}

 HDVs in a scene are simulated along the route centerline by the model in~\cite{strohbeck_deepsil_2021}.
A multiple trajectory prediction network is evaluated every 50ms and the non-colliding prediction with the highest velocity still adhering to traffic rules is selected.
The intersection handling was improved and edge cases were solved to avoid collisions.

The simulated CAVs are based on a single-track vehicle model with parameters taken from our real-world CAV calibration data.
CAVs are controlled by the trajectory planner and controller concept of~\cite{ruof_real-time_2023}, providing realistic CAV behavior.

All HDVs and CAVs do not know the turning directions of other vehicles and therefore have to assume the most conflicting route, except for the set of non-conflicting CAVs explicitly communicated to the CAVs (cf. Section~\ref{ssec:route_knowl}).

\subsubsection{Continuous Traffic Simulation}

Due to limited processing power and quadratically increasing complexity, the simulations are limited to 10 vehicles in total.
We simulate realistic dense traffic by directly reinserting vehicles after they leave the scene.
At a distance of 20\,m after the intersection or roundabout, the vehicles are removed and reinserted into the originating lane.
The reinsertion point is at least 45\,m before the first conflict zone.
If there are vehicles already on the insertion lane, the new vehicle is reinserted further away with the necessary safety distance.
Reinserted vehicles maintain their last speed, but are limited to 30\,km/h.

\subsection{Evaluation Metrics}

We provide efficiency metrics and criticality metrics to compare maneuver planning methods to baselines.

\subsubsection{Efficiency Metrics}

We selected two traffic efficiency metrics and one energy efficiency metric.
Firstly, we calculate the average waiting time in each scenario, that is, the mean duration that vehicles drive below 5\,km/h or stop.
The waiting time should be minimized so that the travel time remains low.
This is basically what our maneuver planning framework directly optimizes.
Next, traffic throughput is considered, that is, the number of vehicles passing the scene during a fixed time.
Higher throughput means that more traffic can pass the scene without causing congestion.
The throughput unit is \si{\per\hour}, that is, vehicles per hour.
Furthermore, the percentage of stopped vehicles is considered, that is, vehicles that reach a speed lower than 1\,km/h.
Fully stopping leads to a loss of energy due to braking, and thus should be prevented.

\subsubsection{Criticality Metrics}

As an increase in traffic efficiency should not be achieved by reducing safety, we also perform a basic criticality analysis.
We consider post-encroachment time (PET) \cite{paul_post_2020} as one of very few metrics that can evaluate the criticality of the situation irrespective of the type of scenario.
The PET between a pair of conflicting vehicles is defined as the duration between the first vehicle leaving and the second vehicle entering the conflict zone.
An encounter is marked critical when the PET of the vehicle pair is lower than the critical threshold of 1\,s (cf.~\cite{paul_post_2020}).

\subsection{Quantitative Results}

\begin{figure}[tb]
	\centering
    \input{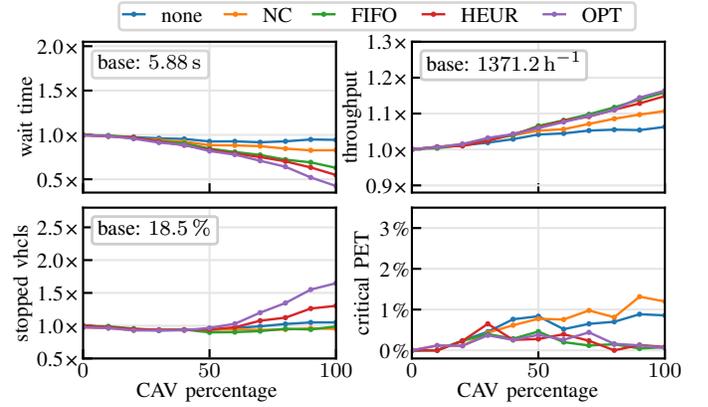}
	\caption{Results of 135 simulations at four intersections with a main road.}
	\label{fig:eval_main_road_inter}
\end{figure}

\begin{figure}[tb]
	\centering
    \input{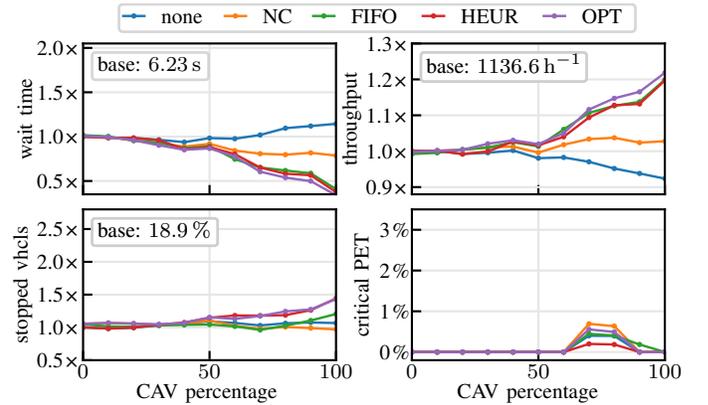}
	\caption{Results of 43 simulated scenarios at one right-before-left intersection.}
	\label{fig:eval_right_bef_left_inter}
\end{figure}

\begin{figure}[tb]
	\centering
    \input{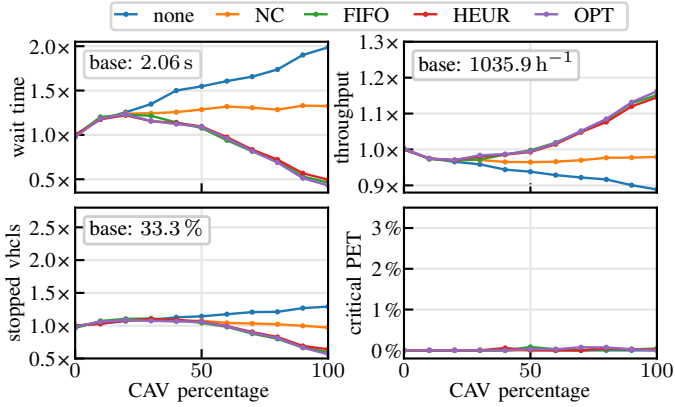}
	\caption{Results of 139 simulated scenarios at seven roundabouts.}
	\label{fig:eval_roundabout}
\end{figure}

\begin{figure}[tb]
	\centering
    \input{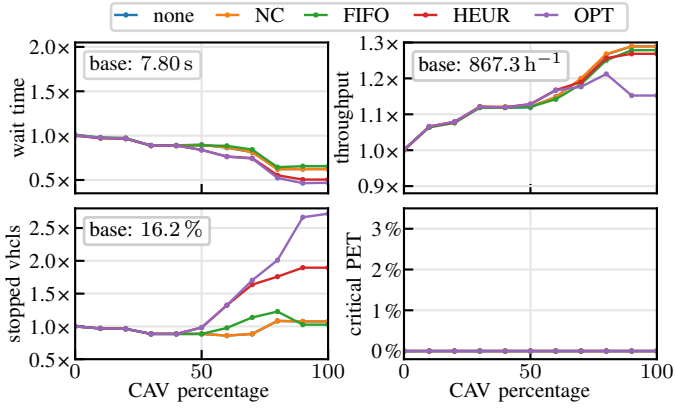}
	\caption{Results of 6 simulated scenarios at a road narrowing.
    }
	\label{fig:eval_narrowing}
\end{figure}

For each CAV percentage and maneuver planning type, we average the metric values over all scenarios.
For easier interpretation, we provide base values and relative numbers for the efficiency metric results.
The results per scenario type are shown in Figs.~\ref{fig:eval_main_road_inter} to~\ref{fig:eval_narrowing}, while the numbers averaged across all scenarios at 40\,\% and 100\,\%~CAVs are given in Tables~\ref{tbl:eff_diff_cav50} and~\ref{tbl:eff_diff_cav100}, respectively.
It should be noted that the results with CAVs without maneuver planning can deviate from the 0\,\%~CAVs results, because the HDV simulation model and the CAV trajectory planning module can exhibit a very different gap acceptance and acceleration behavior.

\begin{table}[t]
	\caption{Metrics results of different maneuver planning methods, averaged over all scenarios, at 40\,\%~CAVs.}
	\label{tbl:eff_diff_cav50}
	\begin{center}
		\begin{tabular}{l r r r r}
			\toprule
			\thead[l]{method} & \thead[r]{wait time} & \thead[r]{throughput} & 
                \thead[r]{vhcl stops} & \thead[r]{critical PET} \\
			\midrule
none & $\SI{4.58}{\second}$ & $\SI{1180.1}{\per\hour}$ & $\SI{26.4}{\percent}$ & $\SI{0.2}{\percent}$ \\
NC & $\SI{0.93}{\times}$ & $\SI{1.01}{\times}$ & \bfseries$\SI{0.97}{\times}$ & $\SI{0.1}{\percent}$ \\
FIFO & $\SI{0.89}{\times}$ & $\SI{1.02}{\times}$ & \bfseries$\SI{0.97}{\times}$ & $\SI{0.1}{\percent}$ \\
HEUR & $\SI{0.88}{\times}$ & \bfseries$\SI{1.03}{\times}$ & $\SI{0.98}{\times}$ & $\SI{0.1}{\percent}$ \\
OPT & \bfseries$\SI{0.87}{\times}$ & \bfseries$\SI{1.03}{\times}$ & \bfseries$\SI{0.97}{\times}$ & $\SI{0.1}{\percent}$ \\
			\bottomrule
		\end{tabular}
	\end{center}
\end{table}

\begin{table}[t]
	\caption{Metrics results of different maneuver planning methods, averaged over all scenarios, at 100\,\%~CAVs.}
	\label{tbl:eff_diff_cav100}
	\begin{center}
		\begin{tabular}{l r r r r}
			\toprule
			\thead[l]{method} & \thead[r]{wait time} & \thead[r]{throughput} & 
                \thead[r]{vhcl stops} & \thead[r]{critical PET} \\
			\midrule
none & $\SI{5.12}{\second}$ & $\SI{1165.9}{\per\hour}$ & $\SI{29.7}{\percent}$ & $\SI{0.3}{\percent}$ \\
NC & $\SI{0.77}{\times}$ & $\SI{1.07}{\times}$ & $\SI{0.81}{\times}$ & $\SI{0.4}{\percent}$ \\
FIFO & $\SI{0.47}{\times}$ & $\SI{1.18}{\times}$ & \bfseries$\SI{0.66}{\times}$ & $\SI{0.1}{\percent}$ \\
HEUR & $\SI{0.42}{\times}$ & $\SI{1.18}{\times}$ & $\SI{0.79}{\times}$ & $\SI{0.0}{\percent}$ \\
OPT & \bfseries$\SI{0.35}{\times}$ & \bfseries$\SI{1.19}{\times}$ & $\SI{0.86}{\times}$ & $\SI{0.0}{\percent}$ \\
			\bottomrule
		\end{tabular}
	\end{center}
\end{table}

\subsubsection{Efficiency Analysis}

Across all intersection and roundabout scenarios, the maneuver planning methods improve the traffic efficiency, i.e., throughput significantly increases and waiting time decreases.
As the OPT method generates the most combinations trying to minimize the overall time loss, naturally, this method achieves the best waiting time reduction, especially at high CAV percentages.
The OPT method also shows the highest increase in throughput when averaged across all scenarios.
Meanwhile, the HEUR approach has slightly lower results, followed by the FIFO baseline.
The knowledge of non-conflicting vehicles baseline already provides a notable efficiency increase.
The methods perform well in mixed traffic and efficiency gains, such as a reduction in waiting time by more than 10\,\%, can be observed starting at 40\,\%~CAVs.
This is consistent with our earlier results in~\cite{mertens_cooperative_2022}.

Regarding the number of stopped vehicles, the FIFO baseline has the best performance, followed by the HEUR and OPT approaches.
This is because both the FIFO and the HEUR method generate orders based on individual vehicle states, prioritizing vehicles close to the stop line and reducing stops.
Meanwhile, the OPT method minimizes the total time loss by deliberately having some additional vehicles yield.

Surprisingly, at roundabouts, deemed to already be efficient, large gains were achieved.
There, cooperation can increase throughput by up to 30.1\,\% and reduce waiting time by a factor of five across all maneuver planning methods.

At the road narrowing, throughput increases with more CAVs, as they drive faster.
In addition to that, throughput cannot be further increased by any reordering, as continuous traffic on the prioritized lane is already optimal.
However, the waiting time can be distributed between the two lanes by alternating priorities.
This complex optimization is achieved only by the HEUR and OPT methods, which shows the effect of the time-dependent weighting in Eq.~\eqref{eq:time_loss_weight}.
Although this reduces the mean waiting time by approximately 10\,\%, it causes a lower throughput and an increased number of stops.

\subsubsection{Criticality Analysis}

The criticality metrics analysis shows that the maneuver planning methods generally do not increase the number of critical encounters.
At intersections with a main road, the trajectory planning seems to have a more active gap acceptance than the HDV model.
However, with active maneuver planning, the number of critical encounters is reduced again.
Thus, maneuver planning has no negative impact, but even a slightly positive effect on the criticality metric, with no significant differences between the planning methods.
All this, of course, highly depends on the trajectory planning implementation and parametrization.

\subsubsection{Run Time Analysis}

We analyze the maximum run time per scenario of each planning method.
The baselines NC and FIFO as well as the HEUR method all have a maximum run time of less than 50\,ms.
The OPT method can take significantly longer, exceeding the cycle time of 200\,ms at least once in 13.3\,\% of the scenarios.
Thus, for high-load scenarios with many vehicles, the implementation of the OPT method needs to be accelerated, or a fallback to one of the simpler methods should be used to remain real-time feasible.

\subsection{Discussion}

The proposed maneuver planning framework works well for mixed traffic and distinguishable efficiency gains are visible starting at a CAV percentage of 40\,\%.
Due to the approach design, both methods cope with all scenario types.
The OPT method achieves the best waiting time and throughput gains, especially at 100\,\%~CAVs, followed by the HEUR method.

The proposed methods sometimes increase the number of stops in order to reduce the total time loss.
If this is undesired, a different trade-off can be made and stops can be penalized in the maneuver selection metric to mitigate this effect.

\section{Conclusion}
\label{sec:conclusion}

By generalizing the underlying driver model, our cooperative maneuver planning framework for mixed traffic now supports a whole class of merging and crossing scenarios present in urban traffic.
Quantitative simulations provide significant insight using real-world V2X protocols and CAV software.
The new optimization approach handles all scenario types well and provides significant efficiency gains even in only partially cooperative traffic without compromising safety.
Meanwhile, the proposed heuristic method results in only slightly lower gains, surpasses the baselines, and remains real-time feasible even in high-load scenarios.

In future works, we want to support other road users like cyclists and buses to resemble even more realistic urban traffic.

\bibliographystyle{IEEEtran}
\bibliography{IEEEabrv,references}

\begin{IEEEbiography}[{\includegraphics[width=1in,height=1.25in,clip,keepaspectratio]{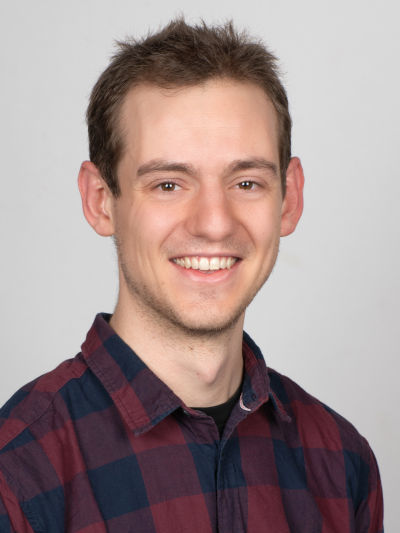}}]{Max Bastian Mertens} earned his M.Sc. degree in Communications and Computer Engineering at Ulm University in 2020. He is a researcher at the Institute of Measurement, Control, and Microtechnology, Ulm University, Ulm, 89081, Germany. His research interests include trajectory and cooperative maneuver planning as well as scene prediction in mixed traffic.
\end{IEEEbiography}

\begin{IEEEbiography}[{\includegraphics[width=1in,height=1.25in,clip,keepaspectratio]{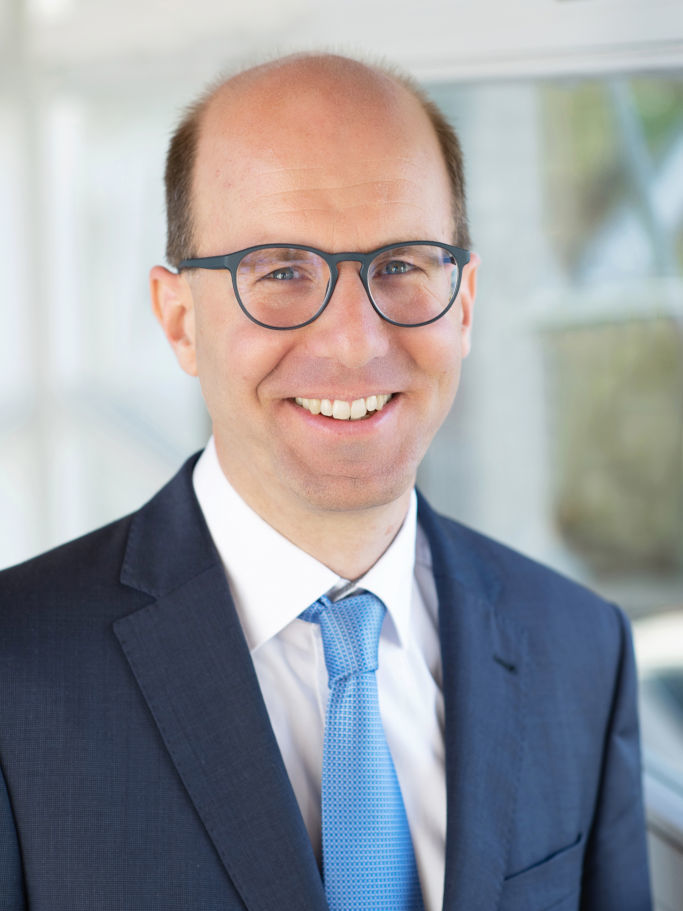}}]{Michael Buchholz} received his Diploma degree in Electrical Engineering and Information Technology as well as his Ph.D. from the faculty of Electrical Engineering and Information Technology at University of Karlsruhe (TH)/Karlsruhe Institute of Technology, Germany.  He is a research group leader and lecturer at the Institute of Measurement, Control, and Microtechnology, Ulm University, Ulm, 89081, Germany, where he earned his ``Habilitation'' (post-doctoral lecturing qualification) for Automation Technology in 2022. His research interests comprise connected automated driving, electric mobility, modelling and control of mechatronic systems, and system identification.
\end{IEEEbiography}

\vfill

\end{document}

%% file: img/scenarios.pdf_tex
%% Creator: Inkscape 1.4 (e7c3feb100, 2024-10-09), www.inkscape.org
%% Accompanies image file 'scenarios.pdf' (pdf, eps, ps)
\begingroup%
  \makeatletter%
  \providecommand\color[2][]{%
    \errmessage{(Inkscape) Color is used for the text in Inkscape, but the package 'color.sty' is not loaded}%
    \renewcommand\color[2][]{}%
  }%
  \providecommand\transparent[1]{%
    \errmessage{(Inkscape) Transparency is used (non-zero) for the text in Inkscape, but the package 'transparent.sty' is not loaded}%
    \renewcommand\transparent[1]{}%
  }%
  \providecommand\rotatebox[2]{#2}%
  \newcommand*\fsize{\dimexpr\f@size pt\relax}%
  \newcommand*\lineheight[1]{\fontsize{\fsize}{#1\fsize}\selectfont}%
  \ifx\svgwidth\undefined%
    \setlength{\unitlength}{581.10417884bp}%
    \ifx\svgscale\undefined%
      \relax%
    \else%
      \setlength{\unitlength}{\unitlength * \real{\svgscale}}%
    \fi%
  \else%
    \setlength{\unitlength}{\svgwidth}%
  \fi%
  \global\let\svgwidth\undefined%
  \global\let\svgscale\undefined%
  \makeatother%
  \begin{picture}(1,0.58371433)%
    \lineheight{1}%
    \setlength\tabcolsep{0pt}%
    \put(0,0){\includegraphics[width=\unitlength,page=1]{scenarios.pdf}}%
    \put(0.23179146,0.2760402){\color[rgb]{0,0,0}\makebox(0,0)[lt]{\lineheight{1.25}\smash{\begin{tabular}[t]{l}(a)\end{tabular}}}}%
    \put(0.23179131,0.00545165){\color[rgb]{0,0,0}\makebox(0,0)[lt]{\lineheight{1.25}\smash{\begin{tabular}[t]{l}(c)\end{tabular}}}}%
    \put(0.74398436,0.00545165){\color[rgb]{0,0,0}\makebox(0,0)[lt]{\lineheight{1.25}\smash{\begin{tabular}[t]{l}(d)\end{tabular}}}}%
    \put(0.74398436,0.27604016){\color[rgb]{0,0,0}\makebox(0,0)[lt]{\lineheight{1.25}\smash{\begin{tabular}[t]{l}(b)\end{tabular}}}}%
  \end{picture}%
\endgroup%

%% file: img/system_overview.pdf_tex
%% Creator: Inkscape 1.4 (e7c3feb100, 2024-10-09), www.inkscape.org
%% Accompanies image file 'system_overview.pdf' (pdf, eps, ps)
\begingroup%
  \makeatletter%
  \providecommand\color[2][]{%
    \errmessage{(Inkscape) Color is used for the text in Inkscape, but the package 'color.sty' is not loaded}%
    \renewcommand\color[2][]{}%
  }%
  \providecommand\transparent[1]{%
    \errmessage{(Inkscape) Transparency is used (non-zero) for the text in Inkscape, but the package 'transparent.sty' is not loaded}%
    \renewcommand\transparent[1]{}%
  }%
  \providecommand\rotatebox[2]{#2}%
  \newcommand*\fsize{\dimexpr\f@size pt\relax}%
  \newcommand*\lineheight[1]{\fontsize{\fsize}{#1\fsize}\selectfont}%
  \ifx\svgwidth\undefined%
    \setlength{\unitlength}{153.92938614bp}%
    \ifx\svgscale\undefined%
      \relax%
    \else%
      \setlength{\unitlength}{\unitlength * \real{\svgscale}}%
    \fi%
  \else%
    \setlength{\unitlength}{\svgwidth}%
  \fi%
  \global\let\svgwidth\undefined%
  \global\let\svgscale\undefined%
  \makeatother%
  \begin{picture}(1,0.44776119)%
    \lineheight{1}%
    \setlength\tabcolsep{0pt}%
    \put(0,0){\includegraphics[width=\unitlength,page=1]{system_overview.pdf}}%
    \put(0.39925375,0.2835821){\makebox(0,0)[lt]{\lineheight{1.25}\smash{\begin{tabular}[t]{l}Maneuver\end{tabular}}}}%
    \put(0.39925375,0.23134329){\makebox(0,0)[lt]{\lineheight{1.25}\smash{\begin{tabular}[t]{l}Planning\end{tabular}}}}%
    \put(0.39925375,0.17537315){\makebox(0,0)[lt]{\lineheight{1.25}\smash{\begin{tabular}[t]{l}Module\end{tabular}}}}%
    \put(0,0){\includegraphics[width=\unitlength,page=2]{system_overview.pdf}}%
    \put(0.09258397,0.15671643){\makebox(0,0)[lt]{\lineheight{1.25}\smash{\begin{tabular}[t]{l}Prediction\end{tabular}}}}%
    \put(0,0){\includegraphics[width=\unitlength,page=3]{system_overview.pdf}}%
    \put(0.10004666,0.04477613){\makebox(0,0)[lt]{\lineheight{1.25}\smash{\begin{tabular}[t]{l}Selection\end{tabular}}}}%
    \put(0,0){\includegraphics[width=\unitlength,page=4]{system_overview.pdf}}%
    \put(0.36561334,0.04477613){\makebox(0,0)[lt]{\lineheight{1.25}\smash{\begin{tabular}[t]{l}Coordination\end{tabular}}}}%
    \put(0,0){\includegraphics[width=\unitlength,page=5]{system_overview.pdf}}%
    \put(0.72014925,0.12313434){\makebox(0,0)[lt]{\lineheight{1.25}\smash{\begin{tabular}[t]{l}Connected\end{tabular}}}}%
    \put(0.72014925,0.07089554){\makebox(0,0)[lt]{\lineheight{1.25}\smash{\begin{tabular}[t]{l}Automated\end{tabular}}}}%
    \put(0.72014925,0.0149254){\makebox(0,0)[lt]{\lineheight{1.25}\smash{\begin{tabular}[t]{l}Vehicles\end{tabular}}}}%
    \put(0,0){\includegraphics[width=\unitlength,page=6]{system_overview.pdf}}%
    \put(0.72014925,0.40298508){\makebox(0,0)[lt]{\lineheight{1.25}\smash{\begin{tabular}[t]{l}Infrastructure\end{tabular}}}}%
    \put(0.72014925,0.34701493){\makebox(0,0)[lt]{\lineheight{1.25}\smash{\begin{tabular}[t]{l}Perception\end{tabular}}}}%
    \put(0,0){\includegraphics[width=\unitlength,page=7]{system_overview.pdf}}%
    \put(0.01492539,0.3955224){\makebox(0,0)[lt]{\lineheight{1.25}\smash{\begin{tabular}[t]{l}Environment Model\end{tabular}}}}%
    \put(0,0){\includegraphics[width=\unitlength,page=8]{system_overview.pdf}}%
    \put(0.08138994,0.26865672){\makebox(0,0)[lt]{\lineheight{1.25}\smash{\begin{tabular}[t]{l}Generation\end{tabular}}}}%
  \end{picture}%
\endgroup%

%% file: img/s_over_t.pgf
%% Creator: Matplotlib, PGF backend
%%
%% To include the figure in your LaTeX document, write
%%   \input{<filename>.pgf}
%%
%% Make sure the required packages are loaded in your preamble
%%   \usepackage{pgf}
%%
%% Also ensure that all the required font packages are loaded; for instance,
%% the lmodern package is sometimes necessary when using math font.
%%   \usepackage{lmodern}
%%
%% Figures using additional raster images can only be included by \input if
%% they are in the same directory as the main LaTeX file. For loading figures
%% from other directories you can use the `import` package
%% and then include the figures with
%%   \import{<path to file>}{<filename>.pgf}
%%
%% Matplotlib used the following preamble
%%   \def\mathdefault#1{#1}
%%   \everymath=\expandafter{\the\everymath\displaystyle}
%%   \usepackage{amsmath}
%%   \ifdefined\pdftexversion\else  % non-pdftex case.
%%     \usepackage{fontspec}
%%     \setmainfont{DejaVuSerif.ttf}[Path=\detokenize{/home/fzt14/.cache/pypoetry/virtualenvs/coop-training-hangPGgN-py3.12/lib/python3.12/site-packages/matplotlib/mpl-data/fonts/ttf/}]
%%     \setsansfont{DejaVuSans.ttf}[Path=\detokenize{/home/fzt14/.cache/pypoetry/virtualenvs/coop-training-hangPGgN-py3.12/lib/python3.12/site-packages/matplotlib/mpl-data/fonts/ttf/}]
%%     \setmonofont{DejaVuSansMono.ttf}[Path=\detokenize{/home/fzt14/.cache/pypoetry/virtualenvs/coop-training-hangPGgN-py3.12/lib/python3.12/site-packages/matplotlib/mpl-data/fonts/ttf/}]
%%   \fi
%%   \makeatletter\@ifpackageloaded{underscore}{}{\usepackage[strings]{underscore}}\makeatother
%%
\begingroup%
\makeatletter%
\begin{pgfpicture}%
\pgfpathrectangle{\pgfpointorigin}{\pgfqpoint{3.700000in}{1.600000in}}%
\pgfusepath{use as bounding box, clip}%
\begin{pgfscope}%
\pgfsetbuttcap%
\pgfsetmiterjoin%
\definecolor{currentfill}{rgb}{1.000000,1.000000,1.000000}%
\pgfsetfillcolor{currentfill}%
\pgfsetlinewidth{0.000000pt}%
\definecolor{currentstroke}{rgb}{1.000000,1.000000,1.000000}%
\pgfsetstrokecolor{currentstroke}%
\pgfsetdash{}{0pt}%
\pgfpathmoveto{\pgfqpoint{0.000000in}{0.000000in}}%
\pgfpathlineto{\pgfqpoint{3.700000in}{0.000000in}}%
\pgfpathlineto{\pgfqpoint{3.700000in}{1.600000in}}%
\pgfpathlineto{\pgfqpoint{0.000000in}{1.600000in}}%
\pgfpathlineto{\pgfqpoint{0.000000in}{0.000000in}}%
\pgfpathclose%
\pgfusepath{fill}%
\end{pgfscope}%
\begin{pgfscope}%
\pgfsetbuttcap%
\pgfsetmiterjoin%
\definecolor{currentfill}{rgb}{1.000000,1.000000,1.000000}%
\pgfsetfillcolor{currentfill}%
\pgfsetlinewidth{0.000000pt}%
\definecolor{currentstroke}{rgb}{0.000000,0.000000,0.000000}%
\pgfsetstrokecolor{currentstroke}%
\pgfsetstrokeopacity{0.000000}%
\pgfsetdash{}{0pt}%
\pgfpathmoveto{\pgfqpoint{0.407000in}{0.400000in}}%
\pgfpathlineto{\pgfqpoint{3.330000in}{0.400000in}}%
\pgfpathlineto{\pgfqpoint{3.330000in}{1.520000in}}%
\pgfpathlineto{\pgfqpoint{0.407000in}{1.520000in}}%
\pgfpathlineto{\pgfqpoint{0.407000in}{0.400000in}}%
\pgfpathclose%
\pgfusepath{fill}%
\end{pgfscope}%
\begin{pgfscope}%
\pgfpathrectangle{\pgfqpoint{0.407000in}{0.400000in}}{\pgfqpoint{2.923000in}{1.120000in}}%
\pgfusepath{clip}%
\pgfsetrectcap%
\pgfsetroundjoin%
\pgfsetlinewidth{0.803000pt}%
\definecolor{currentstroke}{rgb}{0.690196,0.690196,0.690196}%
\pgfsetstrokecolor{currentstroke}%
\pgfsetdash{}{0pt}%
\pgfpathmoveto{\pgfqpoint{0.407000in}{0.400000in}}%
\pgfpathlineto{\pgfqpoint{0.407000in}{1.520000in}}%
\pgfusepath{stroke}%
\end{pgfscope}%
\begin{pgfscope}%
\pgfsetbuttcap%
\pgfsetroundjoin%
\definecolor{currentfill}{rgb}{0.000000,0.000000,0.000000}%
\pgfsetfillcolor{currentfill}%
\pgfsetlinewidth{0.803000pt}%
\definecolor{currentstroke}{rgb}{0.000000,0.000000,0.000000}%
\pgfsetstrokecolor{currentstroke}%
\pgfsetdash{}{0pt}%
\pgfsys@defobject{currentmarker}{\pgfqpoint{0.000000in}{-0.048611in}}{\pgfqpoint{0.000000in}{0.000000in}}{%
\pgfpathmoveto{\pgfqpoint{0.000000in}{0.000000in}}%
\pgfpathlineto{\pgfqpoint{0.000000in}{-0.048611in}}%
\pgfusepath{stroke,fill}%
}%
\begin{pgfscope}%
\pgfsys@transformshift{0.407000in}{0.400000in}%
\pgfsys@useobject{currentmarker}{}%
\end{pgfscope}%
\end{pgfscope}%
\begin{pgfscope}%
\definecolor{textcolor}{rgb}{0.000000,0.000000,0.000000}%
\pgfsetstrokecolor{textcolor}%
\pgfsetfillcolor{textcolor}%
\pgftext[x=0.407000in,y=0.302778in,,top]{\color{textcolor}{\rmfamily\fontsize{8.000000}{9.600000}\selectfont\catcode`\^=\active\def^{\ifmmode\sp\else\^{}\fi}\catcode`\%=\active\def%{\%}$\mathdefault{0}$}}%
\end{pgfscope}%
\begin{pgfscope}%
\pgfpathrectangle{\pgfqpoint{0.407000in}{0.400000in}}{\pgfqpoint{2.923000in}{1.120000in}}%
\pgfusepath{clip}%
\pgfsetrectcap%
\pgfsetroundjoin%
\pgfsetlinewidth{0.803000pt}%
\definecolor{currentstroke}{rgb}{0.690196,0.690196,0.690196}%
\pgfsetstrokecolor{currentstroke}%
\pgfsetdash{}{0pt}%
\pgfpathmoveto{\pgfqpoint{0.991600in}{0.400000in}}%
\pgfpathlineto{\pgfqpoint{0.991600in}{1.520000in}}%
\pgfusepath{stroke}%
\end{pgfscope}%
\begin{pgfscope}%
\pgfsetbuttcap%
\pgfsetroundjoin%
\definecolor{currentfill}{rgb}{0.000000,0.000000,0.000000}%
\pgfsetfillcolor{currentfill}%
\pgfsetlinewidth{0.803000pt}%
\definecolor{currentstroke}{rgb}{0.000000,0.000000,0.000000}%
\pgfsetstrokecolor{currentstroke}%
\pgfsetdash{}{0pt}%
\pgfsys@defobject{currentmarker}{\pgfqpoint{0.000000in}{-0.048611in}}{\pgfqpoint{0.000000in}{0.000000in}}{%
\pgfpathmoveto{\pgfqpoint{0.000000in}{0.000000in}}%
\pgfpathlineto{\pgfqpoint{0.000000in}{-0.048611in}}%
\pgfusepath{stroke,fill}%
}%
\begin{pgfscope}%
\pgfsys@transformshift{0.991600in}{0.400000in}%
\pgfsys@useobject{currentmarker}{}%
\end{pgfscope}%
\end{pgfscope}%
\begin{pgfscope}%
\definecolor{textcolor}{rgb}{0.000000,0.000000,0.000000}%
\pgfsetstrokecolor{textcolor}%
\pgfsetfillcolor{textcolor}%
\pgftext[x=0.991600in,y=0.302778in,,top]{\color{textcolor}{\rmfamily\fontsize{8.000000}{9.600000}\selectfont\catcode`\^=\active\def^{\ifmmode\sp\else\^{}\fi}\catcode`\%=\active\def%{\%}$\mathdefault{2}$}}%
\end{pgfscope}%
\begin{pgfscope}%
\pgfpathrectangle{\pgfqpoint{0.407000in}{0.400000in}}{\pgfqpoint{2.923000in}{1.120000in}}%
\pgfusepath{clip}%
\pgfsetrectcap%
\pgfsetroundjoin%
\pgfsetlinewidth{0.803000pt}%
\definecolor{currentstroke}{rgb}{0.690196,0.690196,0.690196}%
\pgfsetstrokecolor{currentstroke}%
\pgfsetdash{}{0pt}%
\pgfpathmoveto{\pgfqpoint{1.576200in}{0.400000in}}%
\pgfpathlineto{\pgfqpoint{1.576200in}{1.520000in}}%
\pgfusepath{stroke}%
\end{pgfscope}%
\begin{pgfscope}%
\pgfsetbuttcap%
\pgfsetroundjoin%
\definecolor{currentfill}{rgb}{0.000000,0.000000,0.000000}%
\pgfsetfillcolor{currentfill}%
\pgfsetlinewidth{0.803000pt}%
\definecolor{currentstroke}{rgb}{0.000000,0.000000,0.000000}%
\pgfsetstrokecolor{currentstroke}%
\pgfsetdash{}{0pt}%
\pgfsys@defobject{currentmarker}{\pgfqpoint{0.000000in}{-0.048611in}}{\pgfqpoint{0.000000in}{0.000000in}}{%
\pgfpathmoveto{\pgfqpoint{0.000000in}{0.000000in}}%
\pgfpathlineto{\pgfqpoint{0.000000in}{-0.048611in}}%
\pgfusepath{stroke,fill}%
}%
\begin{pgfscope}%
\pgfsys@transformshift{1.576200in}{0.400000in}%
\pgfsys@useobject{currentmarker}{}%
\end{pgfscope}%
\end{pgfscope}%
\begin{pgfscope}%
\definecolor{textcolor}{rgb}{0.000000,0.000000,0.000000}%
\pgfsetstrokecolor{textcolor}%
\pgfsetfillcolor{textcolor}%
\pgftext[x=1.576200in,y=0.302778in,,top]{\color{textcolor}{\rmfamily\fontsize{8.000000}{9.600000}\selectfont\catcode`\^=\active\def^{\ifmmode\sp\else\^{}\fi}\catcode`\%=\active\def%{\%}$\mathdefault{4}$}}%
\end{pgfscope}%
\begin{pgfscope}%
\pgfpathrectangle{\pgfqpoint{0.407000in}{0.400000in}}{\pgfqpoint{2.923000in}{1.120000in}}%
\pgfusepath{clip}%
\pgfsetrectcap%
\pgfsetroundjoin%
\pgfsetlinewidth{0.803000pt}%
\definecolor{currentstroke}{rgb}{0.690196,0.690196,0.690196}%
\pgfsetstrokecolor{currentstroke}%
\pgfsetdash{}{0pt}%
\pgfpathmoveto{\pgfqpoint{2.160800in}{0.400000in}}%
\pgfpathlineto{\pgfqpoint{2.160800in}{1.520000in}}%
\pgfusepath{stroke}%
\end{pgfscope}%
\begin{pgfscope}%
\pgfsetbuttcap%
\pgfsetroundjoin%
\definecolor{currentfill}{rgb}{0.000000,0.000000,0.000000}%
\pgfsetfillcolor{currentfill}%
\pgfsetlinewidth{0.803000pt}%
\definecolor{currentstroke}{rgb}{0.000000,0.000000,0.000000}%
\pgfsetstrokecolor{currentstroke}%
\pgfsetdash{}{0pt}%
\pgfsys@defobject{currentmarker}{\pgfqpoint{0.000000in}{-0.048611in}}{\pgfqpoint{0.000000in}{0.000000in}}{%
\pgfpathmoveto{\pgfqpoint{0.000000in}{0.000000in}}%
\pgfpathlineto{\pgfqpoint{0.000000in}{-0.048611in}}%
\pgfusepath{stroke,fill}%
}%
\begin{pgfscope}%
\pgfsys@transformshift{2.160800in}{0.400000in}%
\pgfsys@useobject{currentmarker}{}%
\end{pgfscope}%
\end{pgfscope}%
\begin{pgfscope}%
\definecolor{textcolor}{rgb}{0.000000,0.000000,0.000000}%
\pgfsetstrokecolor{textcolor}%
\pgfsetfillcolor{textcolor}%
\pgftext[x=2.160800in,y=0.302778in,,top]{\color{textcolor}{\rmfamily\fontsize{8.000000}{9.600000}\selectfont\catcode`\^=\active\def^{\ifmmode\sp\else\^{}\fi}\catcode`\%=\active\def%{\%}$\mathdefault{6}$}}%
\end{pgfscope}%
\begin{pgfscope}%
\pgfpathrectangle{\pgfqpoint{0.407000in}{0.400000in}}{\pgfqpoint{2.923000in}{1.120000in}}%
\pgfusepath{clip}%
\pgfsetrectcap%
\pgfsetroundjoin%
\pgfsetlinewidth{0.803000pt}%
\definecolor{currentstroke}{rgb}{0.690196,0.690196,0.690196}%
\pgfsetstrokecolor{currentstroke}%
\pgfsetdash{}{0pt}%
\pgfpathmoveto{\pgfqpoint{2.745400in}{0.400000in}}%
\pgfpathlineto{\pgfqpoint{2.745400in}{1.520000in}}%
\pgfusepath{stroke}%
\end{pgfscope}%
\begin{pgfscope}%
\pgfsetbuttcap%
\pgfsetroundjoin%
\definecolor{currentfill}{rgb}{0.000000,0.000000,0.000000}%
\pgfsetfillcolor{currentfill}%
\pgfsetlinewidth{0.803000pt}%
\definecolor{currentstroke}{rgb}{0.000000,0.000000,0.000000}%
\pgfsetstrokecolor{currentstroke}%
\pgfsetdash{}{0pt}%
\pgfsys@defobject{currentmarker}{\pgfqpoint{0.000000in}{-0.048611in}}{\pgfqpoint{0.000000in}{0.000000in}}{%
\pgfpathmoveto{\pgfqpoint{0.000000in}{0.000000in}}%
\pgfpathlineto{\pgfqpoint{0.000000in}{-0.048611in}}%
\pgfusepath{stroke,fill}%
}%
\begin{pgfscope}%
\pgfsys@transformshift{2.745400in}{0.400000in}%
\pgfsys@useobject{currentmarker}{}%
\end{pgfscope}%
\end{pgfscope}%
\begin{pgfscope}%
\definecolor{textcolor}{rgb}{0.000000,0.000000,0.000000}%
\pgfsetstrokecolor{textcolor}%
\pgfsetfillcolor{textcolor}%
\pgftext[x=2.745400in,y=0.302778in,,top]{\color{textcolor}{\rmfamily\fontsize{8.000000}{9.600000}\selectfont\catcode`\^=\active\def^{\ifmmode\sp\else\^{}\fi}\catcode`\%=\active\def%{\%}$\mathdefault{8}$}}%
\end{pgfscope}%
\begin{pgfscope}%
\pgfpathrectangle{\pgfqpoint{0.407000in}{0.400000in}}{\pgfqpoint{2.923000in}{1.120000in}}%
\pgfusepath{clip}%
\pgfsetrectcap%
\pgfsetroundjoin%
\pgfsetlinewidth{0.803000pt}%
\definecolor{currentstroke}{rgb}{0.690196,0.690196,0.690196}%
\pgfsetstrokecolor{currentstroke}%
\pgfsetdash{}{0pt}%
\pgfpathmoveto{\pgfqpoint{3.330000in}{0.400000in}}%
\pgfpathlineto{\pgfqpoint{3.330000in}{1.520000in}}%
\pgfusepath{stroke}%
\end{pgfscope}%
\begin{pgfscope}%
\pgfsetbuttcap%
\pgfsetroundjoin%
\definecolor{currentfill}{rgb}{0.000000,0.000000,0.000000}%
\pgfsetfillcolor{currentfill}%
\pgfsetlinewidth{0.803000pt}%
\definecolor{currentstroke}{rgb}{0.000000,0.000000,0.000000}%
\pgfsetstrokecolor{currentstroke}%
\pgfsetdash{}{0pt}%
\pgfsys@defobject{currentmarker}{\pgfqpoint{0.000000in}{-0.048611in}}{\pgfqpoint{0.000000in}{0.000000in}}{%
\pgfpathmoveto{\pgfqpoint{0.000000in}{0.000000in}}%
\pgfpathlineto{\pgfqpoint{0.000000in}{-0.048611in}}%
\pgfusepath{stroke,fill}%
}%
\begin{pgfscope}%
\pgfsys@transformshift{3.330000in}{0.400000in}%
\pgfsys@useobject{currentmarker}{}%
\end{pgfscope}%
\end{pgfscope}%
\begin{pgfscope}%
\definecolor{textcolor}{rgb}{0.000000,0.000000,0.000000}%
\pgfsetstrokecolor{textcolor}%
\pgfsetfillcolor{textcolor}%
\pgftext[x=3.330000in,y=0.302778in,,top]{\color{textcolor}{\rmfamily\fontsize{8.000000}{9.600000}\selectfont\catcode`\^=\active\def^{\ifmmode\sp\else\^{}\fi}\catcode`\%=\active\def%{\%}$\mathdefault{10}$}}%
\end{pgfscope}%
\begin{pgfscope}%
\definecolor{textcolor}{rgb}{0.000000,0.000000,0.000000}%
\pgfsetstrokecolor{textcolor}%
\pgfsetfillcolor{textcolor}%
\pgftext[x=1.868500in,y=0.148457in,,top]{\color{textcolor}{\rmfamily\fontsize{8.000000}{9.600000}\selectfont\catcode`\^=\active\def^{\ifmmode\sp\else\^{}\fi}\catcode`\%=\active\def%{\%}$\Delta t$ in s}}%
\end{pgfscope}%
\begin{pgfscope}%
\pgfpathrectangle{\pgfqpoint{0.407000in}{0.400000in}}{\pgfqpoint{2.923000in}{1.120000in}}%
\pgfusepath{clip}%
\pgfsetrectcap%
\pgfsetroundjoin%
\pgfsetlinewidth{0.803000pt}%
\definecolor{currentstroke}{rgb}{0.690196,0.690196,0.690196}%
\pgfsetstrokecolor{currentstroke}%
\pgfsetdash{}{0pt}%
\pgfpathmoveto{\pgfqpoint{0.407000in}{0.400000in}}%
\pgfpathlineto{\pgfqpoint{3.330000in}{0.400000in}}%
\pgfusepath{stroke}%
\end{pgfscope}%
\begin{pgfscope}%
\pgfsetbuttcap%
\pgfsetroundjoin%
\definecolor{currentfill}{rgb}{0.000000,0.000000,0.000000}%
\pgfsetfillcolor{currentfill}%
\pgfsetlinewidth{0.803000pt}%
\definecolor{currentstroke}{rgb}{0.000000,0.000000,0.000000}%
\pgfsetstrokecolor{currentstroke}%
\pgfsetdash{}{0pt}%
\pgfsys@defobject{currentmarker}{\pgfqpoint{-0.048611in}{0.000000in}}{\pgfqpoint{-0.000000in}{0.000000in}}{%
\pgfpathmoveto{\pgfqpoint{-0.000000in}{0.000000in}}%
\pgfpathlineto{\pgfqpoint{-0.048611in}{0.000000in}}%
\pgfusepath{stroke,fill}%
}%
\begin{pgfscope}%
\pgfsys@transformshift{0.407000in}{0.400000in}%
\pgfsys@useobject{currentmarker}{}%
\end{pgfscope}%
\end{pgfscope}%
\begin{pgfscope}%
\definecolor{textcolor}{rgb}{0.000000,0.000000,0.000000}%
\pgfsetstrokecolor{textcolor}%
\pgfsetfillcolor{textcolor}%
\pgftext[x=0.250749in, y=0.361420in, left, base]{\color{textcolor}{\rmfamily\fontsize{8.000000}{9.600000}\selectfont\catcode`\^=\active\def^{\ifmmode\sp\else\^{}\fi}\catcode`\%=\active\def%{\%}$\mathdefault{0}$}}%
\end{pgfscope}%
\begin{pgfscope}%
\pgfpathrectangle{\pgfqpoint{0.407000in}{0.400000in}}{\pgfqpoint{2.923000in}{1.120000in}}%
\pgfusepath{clip}%
\pgfsetrectcap%
\pgfsetroundjoin%
\pgfsetlinewidth{0.803000pt}%
\definecolor{currentstroke}{rgb}{0.690196,0.690196,0.690196}%
\pgfsetstrokecolor{currentstroke}%
\pgfsetdash{}{0pt}%
\pgfpathmoveto{\pgfqpoint{0.407000in}{0.773333in}}%
\pgfpathlineto{\pgfqpoint{3.330000in}{0.773333in}}%
\pgfusepath{stroke}%
\end{pgfscope}%
\begin{pgfscope}%
\pgfsetbuttcap%
\pgfsetroundjoin%
\definecolor{currentfill}{rgb}{0.000000,0.000000,0.000000}%
\pgfsetfillcolor{currentfill}%
\pgfsetlinewidth{0.803000pt}%
\definecolor{currentstroke}{rgb}{0.000000,0.000000,0.000000}%
\pgfsetstrokecolor{currentstroke}%
\pgfsetdash{}{0pt}%
\pgfsys@defobject{currentmarker}{\pgfqpoint{-0.048611in}{0.000000in}}{\pgfqpoint{-0.000000in}{0.000000in}}{%
\pgfpathmoveto{\pgfqpoint{-0.000000in}{0.000000in}}%
\pgfpathlineto{\pgfqpoint{-0.048611in}{0.000000in}}%
\pgfusepath{stroke,fill}%
}%
\begin{pgfscope}%
\pgfsys@transformshift{0.407000in}{0.773333in}%
\pgfsys@useobject{currentmarker}{}%
\end{pgfscope}%
\end{pgfscope}%
\begin{pgfscope}%
\definecolor{textcolor}{rgb}{0.000000,0.000000,0.000000}%
\pgfsetstrokecolor{textcolor}%
\pgfsetfillcolor{textcolor}%
\pgftext[x=0.191721in, y=0.734753in, left, base]{\color{textcolor}{\rmfamily\fontsize{8.000000}{9.600000}\selectfont\catcode`\^=\active\def^{\ifmmode\sp\else\^{}\fi}\catcode`\%=\active\def%{\%}$\mathdefault{10}$}}%
\end{pgfscope}%
\begin{pgfscope}%
\pgfpathrectangle{\pgfqpoint{0.407000in}{0.400000in}}{\pgfqpoint{2.923000in}{1.120000in}}%
\pgfusepath{clip}%
\pgfsetrectcap%
\pgfsetroundjoin%
\pgfsetlinewidth{0.803000pt}%
\definecolor{currentstroke}{rgb}{0.690196,0.690196,0.690196}%
\pgfsetstrokecolor{currentstroke}%
\pgfsetdash{}{0pt}%
\pgfpathmoveto{\pgfqpoint{0.407000in}{1.146667in}}%
\pgfpathlineto{\pgfqpoint{3.330000in}{1.146667in}}%
\pgfusepath{stroke}%
\end{pgfscope}%
\begin{pgfscope}%
\pgfsetbuttcap%
\pgfsetroundjoin%
\definecolor{currentfill}{rgb}{0.000000,0.000000,0.000000}%
\pgfsetfillcolor{currentfill}%
\pgfsetlinewidth{0.803000pt}%
\definecolor{currentstroke}{rgb}{0.000000,0.000000,0.000000}%
\pgfsetstrokecolor{currentstroke}%
\pgfsetdash{}{0pt}%
\pgfsys@defobject{currentmarker}{\pgfqpoint{-0.048611in}{0.000000in}}{\pgfqpoint{-0.000000in}{0.000000in}}{%
\pgfpathmoveto{\pgfqpoint{-0.000000in}{0.000000in}}%
\pgfpathlineto{\pgfqpoint{-0.048611in}{0.000000in}}%
\pgfusepath{stroke,fill}%
}%
\begin{pgfscope}%
\pgfsys@transformshift{0.407000in}{1.146667in}%
\pgfsys@useobject{currentmarker}{}%
\end{pgfscope}%
\end{pgfscope}%
\begin{pgfscope}%
\definecolor{textcolor}{rgb}{0.000000,0.000000,0.000000}%
\pgfsetstrokecolor{textcolor}%
\pgfsetfillcolor{textcolor}%
\pgftext[x=0.191721in, y=1.108086in, left, base]{\color{textcolor}{\rmfamily\fontsize{8.000000}{9.600000}\selectfont\catcode`\^=\active\def^{\ifmmode\sp\else\^{}\fi}\catcode`\%=\active\def%{\%}$\mathdefault{20}$}}%
\end{pgfscope}%
\begin{pgfscope}%
\pgfpathrectangle{\pgfqpoint{0.407000in}{0.400000in}}{\pgfqpoint{2.923000in}{1.120000in}}%
\pgfusepath{clip}%
\pgfsetrectcap%
\pgfsetroundjoin%
\pgfsetlinewidth{0.803000pt}%
\definecolor{currentstroke}{rgb}{0.690196,0.690196,0.690196}%
\pgfsetstrokecolor{currentstroke}%
\pgfsetdash{}{0pt}%
\pgfpathmoveto{\pgfqpoint{0.407000in}{1.520000in}}%
\pgfpathlineto{\pgfqpoint{3.330000in}{1.520000in}}%
\pgfusepath{stroke}%
\end{pgfscope}%
\begin{pgfscope}%
\pgfsetbuttcap%
\pgfsetroundjoin%
\definecolor{currentfill}{rgb}{0.000000,0.000000,0.000000}%
\pgfsetfillcolor{currentfill}%
\pgfsetlinewidth{0.803000pt}%
\definecolor{currentstroke}{rgb}{0.000000,0.000000,0.000000}%
\pgfsetstrokecolor{currentstroke}%
\pgfsetdash{}{0pt}%
\pgfsys@defobject{currentmarker}{\pgfqpoint{-0.048611in}{0.000000in}}{\pgfqpoint{-0.000000in}{0.000000in}}{%
\pgfpathmoveto{\pgfqpoint{-0.000000in}{0.000000in}}%
\pgfpathlineto{\pgfqpoint{-0.048611in}{0.000000in}}%
\pgfusepath{stroke,fill}%
}%
\begin{pgfscope}%
\pgfsys@transformshift{0.407000in}{1.520000in}%
\pgfsys@useobject{currentmarker}{}%
\end{pgfscope}%
\end{pgfscope}%
\begin{pgfscope}%
\definecolor{textcolor}{rgb}{0.000000,0.000000,0.000000}%
\pgfsetstrokecolor{textcolor}%
\pgfsetfillcolor{textcolor}%
\pgftext[x=0.191721in, y=1.481420in, left, base]{\color{textcolor}{\rmfamily\fontsize{8.000000}{9.600000}\selectfont\catcode`\^=\active\def^{\ifmmode\sp\else\^{}\fi}\catcode`\%=\active\def%{\%}$\mathdefault{30}$}}%
\end{pgfscope}%
\begin{pgfscope}%
\definecolor{textcolor}{rgb}{0.000000,0.000000,0.000000}%
\pgfsetstrokecolor{textcolor}%
\pgfsetfillcolor{textcolor}%
\pgftext[x=0.136165in,y=0.960000in,,bottom,rotate=90.000000]{\color{textcolor}{\rmfamily\fontsize{8.000000}{9.600000}\selectfont\catcode`\^=\active\def^{\ifmmode\sp\else\^{}\fi}\catcode`\%=\active\def%{\%}RMSE in m}}%
\end{pgfscope}%
\begin{pgfscope}%
\pgfpathrectangle{\pgfqpoint{0.407000in}{0.400000in}}{\pgfqpoint{2.923000in}{1.120000in}}%
\pgfusepath{clip}%
\pgfsetbuttcap%
\pgfsetroundjoin%
\pgfsetlinewidth{1.505625pt}%
\definecolor{currentstroke}{rgb}{0.640000,0.150000,0.220000}%
\pgfsetstrokecolor{currentstroke}%
\pgfsetdash{{5.550000pt}{2.400000pt}}{0.000000pt}%
\pgfpathmoveto{\pgfqpoint{0.407000in}{0.400000in}}%
\pgfpathlineto{\pgfqpoint{0.465460in}{0.408604in}}%
\pgfpathlineto{\pgfqpoint{0.523920in}{0.409874in}}%
\pgfpathlineto{\pgfqpoint{0.582380in}{0.414480in}}%
\pgfpathlineto{\pgfqpoint{0.640840in}{0.422458in}}%
\pgfpathlineto{\pgfqpoint{0.699300in}{0.432875in}}%
\pgfpathlineto{\pgfqpoint{0.757760in}{0.445390in}}%
\pgfpathlineto{\pgfqpoint{0.816220in}{0.459724in}}%
\pgfpathlineto{\pgfqpoint{0.874680in}{0.475532in}}%
\pgfpathlineto{\pgfqpoint{0.933140in}{0.492879in}}%
\pgfpathlineto{\pgfqpoint{0.991600in}{0.511127in}}%
\pgfpathlineto{\pgfqpoint{1.050060in}{0.530259in}}%
\pgfpathlineto{\pgfqpoint{1.108520in}{0.550413in}}%
\pgfpathlineto{\pgfqpoint{1.166980in}{0.570993in}}%
\pgfpathlineto{\pgfqpoint{1.225440in}{0.591428in}}%
\pgfpathlineto{\pgfqpoint{1.283900in}{0.611796in}}%
\pgfpathlineto{\pgfqpoint{1.342360in}{0.631686in}}%
\pgfpathlineto{\pgfqpoint{1.400820in}{0.652106in}}%
\pgfpathlineto{\pgfqpoint{1.459280in}{0.672119in}}%
\pgfpathlineto{\pgfqpoint{1.517740in}{0.692367in}}%
\pgfpathlineto{\pgfqpoint{1.576200in}{0.711820in}}%
\pgfpathlineto{\pgfqpoint{1.634660in}{0.731286in}}%
\pgfpathlineto{\pgfqpoint{1.693120in}{0.752049in}}%
\pgfpathlineto{\pgfqpoint{1.751580in}{0.770244in}}%
\pgfpathlineto{\pgfqpoint{1.810040in}{0.789870in}}%
\pgfpathlineto{\pgfqpoint{1.868500in}{0.809623in}}%
\pgfpathlineto{\pgfqpoint{1.926960in}{0.829066in}}%
\pgfpathlineto{\pgfqpoint{1.985420in}{0.850264in}}%
\pgfpathlineto{\pgfqpoint{2.043880in}{0.869791in}}%
\pgfpathlineto{\pgfqpoint{2.102340in}{0.884222in}}%
\pgfpathlineto{\pgfqpoint{2.160800in}{0.898143in}}%
\pgfpathlineto{\pgfqpoint{2.219260in}{0.909585in}}%
\pgfpathlineto{\pgfqpoint{2.277720in}{0.921324in}}%
\pgfpathlineto{\pgfqpoint{2.336180in}{0.924505in}}%
\pgfpathlineto{\pgfqpoint{2.394640in}{0.929705in}}%
\pgfpathlineto{\pgfqpoint{2.453100in}{0.942892in}}%
\pgfpathlineto{\pgfqpoint{2.511560in}{0.951001in}}%
\pgfpathlineto{\pgfqpoint{2.570020in}{0.963304in}}%
\pgfpathlineto{\pgfqpoint{2.628480in}{0.976121in}}%
\pgfpathlineto{\pgfqpoint{2.686940in}{0.989379in}}%
\pgfpathlineto{\pgfqpoint{2.745400in}{0.996405in}}%
\pgfpathlineto{\pgfqpoint{2.803860in}{1.004145in}}%
\pgfpathlineto{\pgfqpoint{2.862320in}{1.006068in}}%
\pgfpathlineto{\pgfqpoint{2.920780in}{1.006726in}}%
\pgfpathlineto{\pgfqpoint{2.979240in}{1.014809in}}%
\pgfpathlineto{\pgfqpoint{3.037700in}{1.018438in}}%
\pgfpathlineto{\pgfqpoint{3.096160in}{1.027631in}}%
\pgfpathlineto{\pgfqpoint{3.154620in}{1.036038in}}%
\pgfpathlineto{\pgfqpoint{3.213080in}{1.027551in}}%
\pgfpathlineto{\pgfqpoint{3.271540in}{1.045565in}}%
\pgfpathlineto{\pgfqpoint{3.330000in}{1.053220in}}%
\pgfpathlineto{\pgfqpoint{3.340000in}{1.053678in}}%
\pgfusepath{stroke}%
\end{pgfscope}%
\begin{pgfscope}%
\pgfpathrectangle{\pgfqpoint{0.407000in}{0.400000in}}{\pgfqpoint{2.923000in}{1.120000in}}%
\pgfusepath{clip}%
\pgfsetrectcap%
\pgfsetroundjoin%
\pgfsetlinewidth{1.505625pt}%
\definecolor{currentstroke}{rgb}{0.640000,0.150000,0.220000}%
\pgfsetstrokecolor{currentstroke}%
\pgfsetdash{}{0pt}%
\pgfpathmoveto{\pgfqpoint{0.407000in}{0.400000in}}%
\pgfpathlineto{\pgfqpoint{0.465460in}{0.408545in}}%
\pgfpathlineto{\pgfqpoint{0.523920in}{0.409223in}}%
\pgfpathlineto{\pgfqpoint{0.582380in}{0.411972in}}%
\pgfpathlineto{\pgfqpoint{0.640840in}{0.417204in}}%
\pgfpathlineto{\pgfqpoint{0.699300in}{0.424277in}}%
\pgfpathlineto{\pgfqpoint{0.757760in}{0.432991in}}%
\pgfpathlineto{\pgfqpoint{0.816220in}{0.443111in}}%
\pgfpathlineto{\pgfqpoint{0.874680in}{0.454250in}}%
\pgfpathlineto{\pgfqpoint{0.933140in}{0.466673in}}%
\pgfpathlineto{\pgfqpoint{0.991600in}{0.479470in}}%
\pgfpathlineto{\pgfqpoint{1.050060in}{0.493030in}}%
\pgfpathlineto{\pgfqpoint{1.108520in}{0.507340in}}%
\pgfpathlineto{\pgfqpoint{1.166980in}{0.522193in}}%
\pgfpathlineto{\pgfqpoint{1.225440in}{0.537497in}}%
\pgfpathlineto{\pgfqpoint{1.283900in}{0.552978in}}%
\pgfpathlineto{\pgfqpoint{1.342360in}{0.568806in}}%
\pgfpathlineto{\pgfqpoint{1.400820in}{0.585403in}}%
\pgfpathlineto{\pgfqpoint{1.459280in}{0.601822in}}%
\pgfpathlineto{\pgfqpoint{1.517740in}{0.618214in}}%
\pgfpathlineto{\pgfqpoint{1.576200in}{0.634892in}}%
\pgfpathlineto{\pgfqpoint{1.634660in}{0.651990in}}%
\pgfpathlineto{\pgfqpoint{1.693120in}{0.669343in}}%
\pgfpathlineto{\pgfqpoint{1.751580in}{0.687803in}}%
\pgfpathlineto{\pgfqpoint{1.810040in}{0.705952in}}%
\pgfpathlineto{\pgfqpoint{1.868500in}{0.722940in}}%
\pgfpathlineto{\pgfqpoint{1.926960in}{0.740956in}}%
\pgfpathlineto{\pgfqpoint{1.985420in}{0.759845in}}%
\pgfpathlineto{\pgfqpoint{2.043880in}{0.775695in}}%
\pgfpathlineto{\pgfqpoint{2.102340in}{0.792043in}}%
\pgfpathlineto{\pgfqpoint{2.160800in}{0.807349in}}%
\pgfpathlineto{\pgfqpoint{2.219260in}{0.821175in}}%
\pgfpathlineto{\pgfqpoint{2.277720in}{0.838457in}}%
\pgfpathlineto{\pgfqpoint{2.336180in}{0.850652in}}%
\pgfpathlineto{\pgfqpoint{2.394640in}{0.865541in}}%
\pgfpathlineto{\pgfqpoint{2.453100in}{0.878709in}}%
\pgfpathlineto{\pgfqpoint{2.511560in}{0.890517in}}%
\pgfpathlineto{\pgfqpoint{2.570020in}{0.906117in}}%
\pgfpathlineto{\pgfqpoint{2.628480in}{0.915788in}}%
\pgfpathlineto{\pgfqpoint{2.686940in}{0.928436in}}%
\pgfpathlineto{\pgfqpoint{2.745400in}{0.940119in}}%
\pgfpathlineto{\pgfqpoint{2.803860in}{0.946406in}}%
\pgfpathlineto{\pgfqpoint{2.862320in}{0.956915in}}%
\pgfpathlineto{\pgfqpoint{2.920780in}{0.965467in}}%
\pgfpathlineto{\pgfqpoint{2.979240in}{0.976353in}}%
\pgfpathlineto{\pgfqpoint{3.037700in}{0.979547in}}%
\pgfpathlineto{\pgfqpoint{3.096160in}{0.985739in}}%
\pgfpathlineto{\pgfqpoint{3.154620in}{0.998117in}}%
\pgfpathlineto{\pgfqpoint{3.213080in}{1.003822in}}%
\pgfpathlineto{\pgfqpoint{3.271540in}{1.018367in}}%
\pgfpathlineto{\pgfqpoint{3.330000in}{1.015047in}}%
\pgfpathlineto{\pgfqpoint{3.340000in}{1.016557in}}%
\pgfusepath{stroke}%
\end{pgfscope}%
\begin{pgfscope}%
\pgfpathrectangle{\pgfqpoint{0.407000in}{0.400000in}}{\pgfqpoint{2.923000in}{1.120000in}}%
\pgfusepath{clip}%
\pgfsetbuttcap%
\pgfsetroundjoin%
\pgfsetlinewidth{1.505625pt}%
\definecolor{currentstroke}{rgb}{0.600000,0.600000,0.600000}%
\pgfsetstrokecolor{currentstroke}%
\pgfsetdash{{5.550000pt}{2.400000pt}}{0.000000pt}%
\pgfpathmoveto{\pgfqpoint{0.407000in}{0.400000in}}%
\pgfpathlineto{\pgfqpoint{0.465460in}{0.408211in}}%
\pgfpathlineto{\pgfqpoint{0.523920in}{0.410030in}}%
\pgfpathlineto{\pgfqpoint{0.582380in}{0.415115in}}%
\pgfpathlineto{\pgfqpoint{0.640840in}{0.423267in}}%
\pgfpathlineto{\pgfqpoint{0.699300in}{0.433711in}}%
\pgfpathlineto{\pgfqpoint{0.757760in}{0.445894in}}%
\pgfpathlineto{\pgfqpoint{0.816220in}{0.459479in}}%
\pgfpathlineto{\pgfqpoint{0.874680in}{0.474229in}}%
\pgfpathlineto{\pgfqpoint{0.933140in}{0.490095in}}%
\pgfpathlineto{\pgfqpoint{0.991600in}{0.507000in}}%
\pgfpathlineto{\pgfqpoint{1.050060in}{0.524833in}}%
\pgfpathlineto{\pgfqpoint{1.108520in}{0.543553in}}%
\pgfpathlineto{\pgfqpoint{1.166980in}{0.563209in}}%
\pgfpathlineto{\pgfqpoint{1.225440in}{0.583748in}}%
\pgfpathlineto{\pgfqpoint{1.283900in}{0.605061in}}%
\pgfpathlineto{\pgfqpoint{1.342360in}{0.627107in}}%
\pgfpathlineto{\pgfqpoint{1.400820in}{0.649892in}}%
\pgfpathlineto{\pgfqpoint{1.459280in}{0.673416in}}%
\pgfpathlineto{\pgfqpoint{1.517740in}{0.697670in}}%
\pgfpathlineto{\pgfqpoint{1.576200in}{0.722417in}}%
\pgfpathlineto{\pgfqpoint{1.634660in}{0.747814in}}%
\pgfpathlineto{\pgfqpoint{1.693120in}{0.773725in}}%
\pgfpathlineto{\pgfqpoint{1.751580in}{0.800129in}}%
\pgfpathlineto{\pgfqpoint{1.810040in}{0.826967in}}%
\pgfpathlineto{\pgfqpoint{1.868500in}{0.854352in}}%
\pgfpathlineto{\pgfqpoint{1.926960in}{0.881785in}}%
\pgfpathlineto{\pgfqpoint{1.985420in}{0.909654in}}%
\pgfpathlineto{\pgfqpoint{2.043880in}{0.937321in}}%
\pgfpathlineto{\pgfqpoint{2.102340in}{0.965551in}}%
\pgfpathlineto{\pgfqpoint{2.160800in}{0.993768in}}%
\pgfpathlineto{\pgfqpoint{2.219260in}{1.022625in}}%
\pgfpathlineto{\pgfqpoint{2.277720in}{1.051394in}}%
\pgfpathlineto{\pgfqpoint{2.336180in}{1.080204in}}%
\pgfpathlineto{\pgfqpoint{2.394640in}{1.109064in}}%
\pgfpathlineto{\pgfqpoint{2.453100in}{1.138362in}}%
\pgfpathlineto{\pgfqpoint{2.511560in}{1.167003in}}%
\pgfpathlineto{\pgfqpoint{2.570020in}{1.195883in}}%
\pgfpathlineto{\pgfqpoint{2.628480in}{1.225078in}}%
\pgfpathlineto{\pgfqpoint{2.686940in}{1.253897in}}%
\pgfpathlineto{\pgfqpoint{2.745400in}{1.282243in}}%
\pgfpathlineto{\pgfqpoint{2.803860in}{1.310536in}}%
\pgfpathlineto{\pgfqpoint{2.862320in}{1.338515in}}%
\pgfpathlineto{\pgfqpoint{2.920780in}{1.365188in}}%
\pgfpathlineto{\pgfqpoint{2.979240in}{1.392055in}}%
\pgfpathlineto{\pgfqpoint{3.037700in}{1.417763in}}%
\pgfpathlineto{\pgfqpoint{3.096160in}{1.443899in}}%
\pgfpathlineto{\pgfqpoint{3.154620in}{1.469067in}}%
\pgfpathlineto{\pgfqpoint{3.213080in}{1.493055in}}%
\pgfpathlineto{\pgfqpoint{3.271540in}{1.516540in}}%
\pgfpathlineto{\pgfqpoint{3.307869in}{1.530000in}}%
\pgfusepath{stroke}%
\end{pgfscope}%
\begin{pgfscope}%
\pgfpathrectangle{\pgfqpoint{0.407000in}{0.400000in}}{\pgfqpoint{2.923000in}{1.120000in}}%
\pgfusepath{clip}%
\pgfsetrectcap%
\pgfsetroundjoin%
\pgfsetlinewidth{1.505625pt}%
\definecolor{currentstroke}{rgb}{0.600000,0.600000,0.600000}%
\pgfsetstrokecolor{currentstroke}%
\pgfsetdash{}{0pt}%
\pgfpathmoveto{\pgfqpoint{0.407000in}{0.400000in}}%
\pgfpathlineto{\pgfqpoint{0.465460in}{0.408063in}}%
\pgfpathlineto{\pgfqpoint{0.523920in}{0.409295in}}%
\pgfpathlineto{\pgfqpoint{0.582380in}{0.413071in}}%
\pgfpathlineto{\pgfqpoint{0.640840in}{0.419368in}}%
\pgfpathlineto{\pgfqpoint{0.699300in}{0.427568in}}%
\pgfpathlineto{\pgfqpoint{0.757760in}{0.437269in}}%
\pgfpathlineto{\pgfqpoint{0.816220in}{0.448233in}}%
\pgfpathlineto{\pgfqpoint{0.874680in}{0.460236in}}%
\pgfpathlineto{\pgfqpoint{0.933140in}{0.473154in}}%
\pgfpathlineto{\pgfqpoint{0.991600in}{0.486824in}}%
\pgfpathlineto{\pgfqpoint{1.050060in}{0.501098in}}%
\pgfpathlineto{\pgfqpoint{1.108520in}{0.515922in}}%
\pgfpathlineto{\pgfqpoint{1.166980in}{0.531319in}}%
\pgfpathlineto{\pgfqpoint{1.225440in}{0.547159in}}%
\pgfpathlineto{\pgfqpoint{1.283900in}{0.563337in}}%
\pgfpathlineto{\pgfqpoint{1.342360in}{0.579813in}}%
\pgfpathlineto{\pgfqpoint{1.400820in}{0.596585in}}%
\pgfpathlineto{\pgfqpoint{1.459280in}{0.613523in}}%
\pgfpathlineto{\pgfqpoint{1.517740in}{0.630894in}}%
\pgfpathlineto{\pgfqpoint{1.576200in}{0.648208in}}%
\pgfpathlineto{\pgfqpoint{1.634660in}{0.665947in}}%
\pgfpathlineto{\pgfqpoint{1.693120in}{0.683521in}}%
\pgfpathlineto{\pgfqpoint{1.751580in}{0.701009in}}%
\pgfpathlineto{\pgfqpoint{1.810040in}{0.718707in}}%
\pgfpathlineto{\pgfqpoint{1.868500in}{0.736909in}}%
\pgfpathlineto{\pgfqpoint{1.926960in}{0.754928in}}%
\pgfpathlineto{\pgfqpoint{1.985420in}{0.773237in}}%
\pgfpathlineto{\pgfqpoint{2.043880in}{0.790699in}}%
\pgfpathlineto{\pgfqpoint{2.102340in}{0.809103in}}%
\pgfpathlineto{\pgfqpoint{2.160800in}{0.826968in}}%
\pgfpathlineto{\pgfqpoint{2.219260in}{0.845191in}}%
\pgfpathlineto{\pgfqpoint{2.277720in}{0.863760in}}%
\pgfpathlineto{\pgfqpoint{2.336180in}{0.881771in}}%
\pgfpathlineto{\pgfqpoint{2.394640in}{0.899805in}}%
\pgfpathlineto{\pgfqpoint{2.453100in}{0.918635in}}%
\pgfpathlineto{\pgfqpoint{2.511560in}{0.936780in}}%
\pgfpathlineto{\pgfqpoint{2.570020in}{0.955492in}}%
\pgfpathlineto{\pgfqpoint{2.628480in}{0.973203in}}%
\pgfpathlineto{\pgfqpoint{2.686940in}{0.991458in}}%
\pgfpathlineto{\pgfqpoint{2.745400in}{1.009035in}}%
\pgfpathlineto{\pgfqpoint{2.803860in}{1.026491in}}%
\pgfpathlineto{\pgfqpoint{2.862320in}{1.043655in}}%
\pgfpathlineto{\pgfqpoint{2.920780in}{1.060742in}}%
\pgfpathlineto{\pgfqpoint{2.979240in}{1.078885in}}%
\pgfpathlineto{\pgfqpoint{3.037700in}{1.096121in}}%
\pgfpathlineto{\pgfqpoint{3.096160in}{1.112500in}}%
\pgfpathlineto{\pgfqpoint{3.154620in}{1.129533in}}%
\pgfpathlineto{\pgfqpoint{3.213080in}{1.146287in}}%
\pgfpathlineto{\pgfqpoint{3.271540in}{1.163787in}}%
\pgfpathlineto{\pgfqpoint{3.330000in}{1.180485in}}%
\pgfpathlineto{\pgfqpoint{3.340000in}{1.183266in}}%
\pgfusepath{stroke}%
\end{pgfscope}%
\begin{pgfscope}%
\pgfsetrectcap%
\pgfsetmiterjoin%
\pgfsetlinewidth{0.803000pt}%
\definecolor{currentstroke}{rgb}{0.000000,0.000000,0.000000}%
\pgfsetstrokecolor{currentstroke}%
\pgfsetdash{}{0pt}%
\pgfpathmoveto{\pgfqpoint{0.407000in}{0.400000in}}%
\pgfpathlineto{\pgfqpoint{0.407000in}{1.520000in}}%
\pgfusepath{stroke}%
\end{pgfscope}%
\begin{pgfscope}%
\pgfsetrectcap%
\pgfsetmiterjoin%
\pgfsetlinewidth{0.803000pt}%
\definecolor{currentstroke}{rgb}{0.000000,0.000000,0.000000}%
\pgfsetstrokecolor{currentstroke}%
\pgfsetdash{}{0pt}%
\pgfpathmoveto{\pgfqpoint{3.330000in}{0.400000in}}%
\pgfpathlineto{\pgfqpoint{3.330000in}{1.520000in}}%
\pgfusepath{stroke}%
\end{pgfscope}%
\begin{pgfscope}%
\pgfsetrectcap%
\pgfsetmiterjoin%
\pgfsetlinewidth{0.803000pt}%
\definecolor{currentstroke}{rgb}{0.000000,0.000000,0.000000}%
\pgfsetstrokecolor{currentstroke}%
\pgfsetdash{}{0pt}%
\pgfpathmoveto{\pgfqpoint{0.407000in}{0.400000in}}%
\pgfpathlineto{\pgfqpoint{3.330000in}{0.400000in}}%
\pgfusepath{stroke}%
\end{pgfscope}%
\begin{pgfscope}%
\pgfsetrectcap%
\pgfsetmiterjoin%
\pgfsetlinewidth{0.803000pt}%
\definecolor{currentstroke}{rgb}{0.000000,0.000000,0.000000}%
\pgfsetstrokecolor{currentstroke}%
\pgfsetdash{}{0pt}%
\pgfpathmoveto{\pgfqpoint{0.407000in}{1.520000in}}%
\pgfpathlineto{\pgfqpoint{3.330000in}{1.520000in}}%
\pgfusepath{stroke}%
\end{pgfscope}%
\begin{pgfscope}%
\pgfsetbuttcap%
\pgfsetmiterjoin%
\definecolor{currentfill}{rgb}{1.000000,1.000000,1.000000}%
\pgfsetfillcolor{currentfill}%
\pgfsetfillopacity{0.800000}%
\pgfsetlinewidth{1.003750pt}%
\definecolor{currentstroke}{rgb}{0.800000,0.800000,0.800000}%
\pgfsetstrokecolor{currentstroke}%
\pgfsetstrokeopacity{0.800000}%
\pgfsetdash{}{0pt}%
\pgfpathmoveto{\pgfqpoint{0.484778in}{0.811358in}}%
\pgfpathlineto{\pgfqpoint{1.579426in}{0.811358in}}%
\pgfpathquadraticcurveto{\pgfqpoint{1.601648in}{0.811358in}}{\pgfqpoint{1.601648in}{0.833580in}}%
\pgfpathlineto{\pgfqpoint{1.601648in}{1.442222in}}%
\pgfpathquadraticcurveto{\pgfqpoint{1.601648in}{1.464444in}}{\pgfqpoint{1.579426in}{1.464444in}}%
\pgfpathlineto{\pgfqpoint{0.484778in}{1.464444in}}%
\pgfpathquadraticcurveto{\pgfqpoint{0.462556in}{1.464444in}}{\pgfqpoint{0.462556in}{1.442222in}}%
\pgfpathlineto{\pgfqpoint{0.462556in}{0.833580in}}%
\pgfpathquadraticcurveto{\pgfqpoint{0.462556in}{0.811358in}}{\pgfqpoint{0.484778in}{0.811358in}}%
\pgfpathlineto{\pgfqpoint{0.484778in}{0.811358in}}%
\pgfpathclose%
\pgfusepath{stroke,fill}%
\end{pgfscope}%
\begin{pgfscope}%
\pgfsetbuttcap%
\pgfsetroundjoin%
\pgfsetlinewidth{1.505625pt}%
\definecolor{currentstroke}{rgb}{0.640000,0.150000,0.220000}%
\pgfsetstrokecolor{currentstroke}%
\pgfsetdash{{5.550000pt}{2.400000pt}}{0.000000pt}%
\pgfpathmoveto{\pgfqpoint{0.507000in}{1.381111in}}%
\pgfpathlineto{\pgfqpoint{0.618111in}{1.381111in}}%
\pgfpathlineto{\pgfqpoint{0.729222in}{1.381111in}}%
\pgfusepath{stroke}%
\end{pgfscope}%
\begin{pgfscope}%
\definecolor{textcolor}{rgb}{0.000000,0.000000,0.000000}%
\pgfsetstrokecolor{textcolor}%
\pgfsetfillcolor{textcolor}%
\pgftext[x=0.818111in,y=1.342222in,left,base]{\color{textcolor}{\rmfamily\fontsize{8.000000}{9.600000}\selectfont\catcode`\^=\active\def^{\ifmmode\sp\else\^{}\fi}\catcode`\%=\active\def%{\%}inD, IDM}}%
\end{pgfscope}%
\begin{pgfscope}%
\pgfsetrectcap%
\pgfsetroundjoin%
\pgfsetlinewidth{1.505625pt}%
\definecolor{currentstroke}{rgb}{0.640000,0.150000,0.220000}%
\pgfsetstrokecolor{currentstroke}%
\pgfsetdash{}{0pt}%
\pgfpathmoveto{\pgfqpoint{0.507000in}{1.226173in}}%
\pgfpathlineto{\pgfqpoint{0.618111in}{1.226173in}}%
\pgfpathlineto{\pgfqpoint{0.729222in}{1.226173in}}%
\pgfusepath{stroke}%
\end{pgfscope}%
\begin{pgfscope}%
\definecolor{textcolor}{rgb}{0.000000,0.000000,0.000000}%
\pgfsetstrokecolor{textcolor}%
\pgfsetfillcolor{textcolor}%
\pgftext[x=0.818111in,y=1.187284in,left,base]{\color{textcolor}{\rmfamily\fontsize{8.000000}{9.600000}\selectfont\catcode`\^=\active\def^{\ifmmode\sp\else\^{}\fi}\catcode`\%=\active\def%{\%}inD, ours}}%
\end{pgfscope}%
\begin{pgfscope}%
\pgfsetbuttcap%
\pgfsetroundjoin%
\pgfsetlinewidth{1.505625pt}%
\definecolor{currentstroke}{rgb}{0.600000,0.600000,0.600000}%
\pgfsetstrokecolor{currentstroke}%
\pgfsetdash{{5.550000pt}{2.400000pt}}{0.000000pt}%
\pgfpathmoveto{\pgfqpoint{0.507000in}{1.071234in}}%
\pgfpathlineto{\pgfqpoint{0.618111in}{1.071234in}}%
\pgfpathlineto{\pgfqpoint{0.729222in}{1.071234in}}%
\pgfusepath{stroke}%
\end{pgfscope}%
\begin{pgfscope}%
\definecolor{textcolor}{rgb}{0.000000,0.000000,0.000000}%
\pgfsetstrokecolor{textcolor}%
\pgfsetfillcolor{textcolor}%
\pgftext[x=0.818111in,y=1.032346in,left,base]{\color{textcolor}{\rmfamily\fontsize{8.000000}{9.600000}\selectfont\catcode`\^=\active\def^{\ifmmode\sp\else\^{}\fi}\catcode`\%=\active\def%{\%}openDD, IDM}}%
\end{pgfscope}%
\begin{pgfscope}%
\pgfsetrectcap%
\pgfsetroundjoin%
\pgfsetlinewidth{1.505625pt}%
\definecolor{currentstroke}{rgb}{0.600000,0.600000,0.600000}%
\pgfsetstrokecolor{currentstroke}%
\pgfsetdash{}{0pt}%
\pgfpathmoveto{\pgfqpoint{0.507000in}{0.916296in}}%
\pgfpathlineto{\pgfqpoint{0.618111in}{0.916296in}}%
\pgfpathlineto{\pgfqpoint{0.729222in}{0.916296in}}%
\pgfusepath{stroke}%
\end{pgfscope}%
\begin{pgfscope}%
\definecolor{textcolor}{rgb}{0.000000,0.000000,0.000000}%
\pgfsetstrokecolor{textcolor}%
\pgfsetfillcolor{textcolor}%
\pgftext[x=0.818111in,y=0.877407in,left,base]{\color{textcolor}{\rmfamily\fontsize{8.000000}{9.600000}\selectfont\catcode`\^=\active\def^{\ifmmode\sp\else\^{}\fi}\catcode`\%=\active\def%{\%}openDD, ours}}%
\end{pgfscope}%
\end{pgfpicture}%
\makeatother%
\endgroup%

%% file: root.bbl
\begin{thebibliography}{10}
\providecommand{\url}[1]{#1}
\csname url@samestyle\endcsname
\providecommand{\newblock}{\relax}
\providecommand{\bibinfo}[2]{#2}
\providecommand{\BIBentrySTDinterwordspacing}{\spaceskip=0pt\relax}
\providecommand{\BIBentryALTinterwordstretchfactor}{4}
\providecommand{\BIBentryALTinterwordspacing}{\spaceskip=\fontdimen2\font plus
\BIBentryALTinterwordstretchfactor\fontdimen3\font minus \fontdimen4\font\relax}
\providecommand{\BIBforeignlanguage}[2]{{%
\expandafter\ifx\csname l@#1\endcsname\relax
\typeout{** WARNING: IEEEtran.bst: No hyphenation pattern has been}%
\typeout{** loaded for the language `#1'. Using the pattern for}%
\typeout{** the default language instead.}%
\else
\language=\csname l@#1\endcsname
\fi
#2}}
\providecommand{\BIBdecl}{\relax}
\BIBdecl

\bibitem{buchholz_handling_2021}
M.~Buchholz, J.~C. Müller, M.~Herrmann, J.~Strohbeck, B.~Völz, M.~Maier, J.~Paczia, O.~Stein, H.~Rehborn, and R.-W. Henn, ``Handling {Occlusions} in {Automated} {Driving} {Using} a {Multiaccess} {Edge} {Computing} {Server}-{Based} {Environment} {Model} {From} {Infrastructure} {Sensors},'' \emph{IEEE Intell. Transp. Syst. Magazine}, pp. 2--16, 2021.

\bibitem{mertens_cooperative_2022}
M.~B. Mertens, J.~Müller, and M.~Buchholz, ``Cooperative {Maneuver} {Planning} for {Mixed} {Traffic} at {Unsignalized} {Intersections} {Using} {Probabilistic} {Predictions},'' in \emph{2022 {IEEE} {Intell.} {Veh.} {Symp.} ({IV})}, 2022, pp. 1174--1180.

\bibitem{mertens_fast_2024}
M.~B. Mertens, J.~Ruof, J.~Strohbeck, and M.~Buchholz, ``Fast {Long}-{Term} {Multi}-{Scenario} {Prediction} for {Maneuver} {Planning} at {Unsignalized} {Intersections},'' in \emph{2024 {American} {Control} {Conf.} ({ACC})}, 2024, pp. 111--116.

\bibitem{klimke_real-world_2025}
M.~Klimke, M.~B. Mertens, B.~Völz, and M.~Buchholz, ``Real-{World} {Evaluation} of two {Cooperative} {Intersection} {Management} {Approaches},'' 2025, arXiv:2403.16478.

\bibitem{strohbeck_deepsil_2021}
J.~Strohbeck, J.~Müller, A.~Holzbock, and M.~Buchholz, ``{DeepSIL}: {A} {Software}-in-the-{Loop} {Framework} for {Evaluating} {Motion} {Planning} {Schemes} {Using} {Multiple} {Trajectory} {Prediction} {Networks}$^{\textrm{*}}$,'' in \emph{2021 {IEEE}/{RSJ} {Int.} {Conf.} on {Intell.} {Robots} and {Syst.} ({IROS})}, 2021, pp. 7075--7081.

\bibitem{ruof_real-time_2023}
J.~Ruof, M.~B. Mertens, M.~Buchholz, and K.~Dietmayer, ``\BIBforeignlanguage{en}{Real-{Time} {Spatial} {Trajectory} {Planning} for {Urban} {Environments} {Using} {Dynamic} {Optimization}},'' in \emph{\BIBforeignlanguage{en}{2023 {IEEE} {Intell.} {Veh.} {Symp.} ({IV})}}, 2023.

\bibitem{mertens_extended_2021}
M.~B. Mertens, J.~Müller, R.~Dehler, M.~Klimke, M.~Maier, S.~Gherekhloo, B.~Völz, R.-W. Henn, and M.~Buchholz, ``\BIBforeignlanguage{en}{An extended maneuver coordination protocol with support for urban scenarios and mixed traffic},'' in \emph{\BIBforeignlanguage{en}{2021 {IEEE} {Veh.} {Netw.} {Conf.} ({VNC})}}, vol. 2021, 2021, pp. 32--35.

\bibitem{dresner_multiagent_2004}
K.~Dresner and P.~Stone, ``Multiagent traffic management: a reservation-based intersection control mechanism,'' in \emph{Proc. of the {Third} {Int.} {Joint} {Conf.} on {Autonomous} {Agents} and {Multiagent} {Syst.}, 2004. {AAMAS} 2004.}, 2004, pp. 530--537.

\bibitem{zhong_autonomous_2021}
Z.~Zhong, M.~Nejad, and E.~E. Lee, ``Autonomous and {Semiautonomous} {Intersection} {Management}: {A} {Survey},'' \emph{IEEE Intell. Transp. Syst. Magazine}, vol.~13, no.~2, pp. 53--70, 2021.

\bibitem{li_cooperative_2006}
L.~Li and F.-Y. Wang, ``Cooperative {Driving} at {Blind} {Crossings} {Using} {Intervehicle} {Communication},'' \emph{IEEE Transactions on Veh. Technol.}, vol.~55, no.~6, pp. 1712--1724, 2006.

\bibitem{wu_cooperative_2012}
J.~Wu, A.~Abbas-Turki, and A.~El Moudni, ``\BIBforeignlanguage{en}{Cooperative driving: an ant colony system for autonomous intersection management},'' \emph{\BIBforeignlanguage{en}{Appl Intell}}, vol.~37, no.~2, pp. 207--222, 2012.

\bibitem{wuthishuwong_vehicle_2013}
C.~Wuthishuwong and A.~Traechtler, ``Vehicle to infrastructure based safe trajectory planning for {Autonomous} {Intersection} {Management},'' in \emph{2013 13th {Int.} {Conf.} on {ITS} {Telecommun.} ({ITST})}, 2013, pp. 175--180.

\bibitem{ahmane_modeling_2013}
M.~Ahmane, A.~Abbas-Turki, F.~Perronnet, J.~Wu, A.~E. Moudni, J.~Buisson, and R.~Zeo, ``\BIBforeignlanguage{en}{Modeling and controlling an isolated urban intersection based on cooperative vehicles},'' \emph{\BIBforeignlanguage{en}{Transp. Research Part C: Emerging Technologies}}, vol.~28, pp. 44--62, 2013.

\bibitem{kurzer_decentralized_2018}
K.~Kurzer, C.~Zhou, and J.~Marius~Zöllner, ``Decentralized {Cooperative} {Planning} for {Automated} {Vehicles} with {Hierarchical} {Monte} {Carlo} {Tree} {Search},'' in \emph{2018 {IEEE} {Intell.} {Veh.} {Symp.} ({IV})}, 2018, pp. 529--536.

\bibitem{bichiou_developing_2019}
Y.~Bichiou and H.~A. Rakha, ``Developing an {Optimal} {Intersection} {Control} {System} for {Automated} {Connected} {Vehicles},'' \emph{IEEE Transactions on Intell. Transp. Syst.}, vol.~20, no.~5, pp. 1908--1916, 2019.

\bibitem{Martin-Gasulla_traffic_2021}
M.~Martin-Gasulla and L.~Elefteriadou, ``Traffic management with autonomous and connected vehicles at single-lane roundabouts,'' \emph{Transp. Research Part C: Emerging Technologies}, vol. 125, p. 102964, 2021.

\bibitem{bento_intelligent_2012}
L.~C. Bento, R.~Parafita, and U.~Nunes, ``Intelligent traffic management at intersections supported by {V2V} and {V2I} communications,'' in \emph{2012 15th {Int.} {IEEE} {Conf.} on {Intell.} {Transp.} {Syst.}}, 2012, pp. 1495--1502.

\bibitem{dresner_sharing_2007}
K.~Dresner and P.~Stone, ``Sharing the {Road}: {Autonomous} {Vehicles} {Meet} {Human} {Drivers},'' in \emph{{IJCAI}}, 2007.

\bibitem{bento_intelligent_2013}
L.~C. Bento, R.~Parafita, S.~Santos, and U.~Nunes, ``Intelligent traffic management at intersections: {Legacy} mode for vehicles not equipped with {V2V} and {V2I} communications,'' in \emph{16th {Int.} {IEEE} {Conf.} on {Intell.} {Transp.} {Syst.} ({ITSC} 2013)}, 2013, pp. 726--731.

\bibitem{qian_priority-based_2014}
X.~Qian, J.~Gregoire, F.~Moutarde, and A.~De~La~Fortelle, ``Priority-based coordination of autonomous and legacy vehicles at intersection,'' in \emph{17th {Int.} {IEEE} {Conf.} on {Intell.} {Transp.} {Syst.} ({ITSC})}, 2014, pp. 1166--1171.

\bibitem{yang_isolated_2016}
K.~Yang, S.~I. Guler, and M.~Menendez, ``\BIBforeignlanguage{en}{Isolated intersection control for various levels of vehicle technology: {Conventional}, connected, and automated vehicles},'' \emph{\BIBforeignlanguage{en}{Transp. Research Part C: Emerging Technologies}}, vol.~72, pp. 109--129, 2016.

\bibitem{zhao_optimal_2018}
L.~Zhao, A.~Malikopoulos, and J.~Rios-Torres, ``Optimal {Control} of {Connected} and {Automated} {Vehicles} at {Roundabouts}: {An} {Investigation} in a {Mixed}-{Traffic} {Environment}⁎,'' \emph{IFAC-PapersOnLine}, vol.~51, no.~9, pp. 73--78, 2018.

\bibitem{nichting_space_2020}
M.~Nichting, D.~Heß, J.~Schindler, T.~Hesse, and F.~Köster, ``Space {Time} {Reservation} {Procedure} ({STRP}) for {V2X}-{Based} {Maneuver} {Coordination} of {Cooperative} {Automated} {Vehicles} in {Diverse} {Conflict} {Scenarios},'' in \emph{2020 {IEEE} {Intell.} {Veh.} {Symp.} ({IV})}, 2020, pp. 502--509.

\bibitem{klimke_automatic_2023}
M.~Klimke, B.~Völz, and M.~Buchholz, ``Automatic {Intersection} {Management} in {Mixed} {Traffic} {Using} {Reinforcement} {Learning} and {Graph} {Neural} {Networks},'' in \emph{2023 {IEEE} {Intell.} {Veh.} {Symp.} ({IV})}, 2023, pp. 1--8.

\bibitem{bock_ind_2020}
J.~Bock, R.~Krajewski, T.~Moers, S.~Runde, L.~Vater, and L.~Eckstein, ``The {inD} {Dataset}: {A} {Drone} {Dataset} of {Naturalistic} {Road} {User} {Trajectories} at {German} {Intersections},'' in \emph{2020 {IEEE} {Intell.} {Veh.} {Symp.} ({IV})}, 2020, pp. 1929--1934.

\bibitem{breuer_opendd_2020}
A.~Breuer, J.-A. Termöhlen, S.~Homoceanu, and T.~Fingscheidt, ``{openDD}: {A} {Large}-{Scale} {Roundabout} {Drone} {Dataset},'' 2020, arXiv:2007.08463.

\bibitem{schulman_proximal_2017}
J.~Schulman, F.~Wolski, P.~Dhariwal, A.~Radford, and O.~Klimov, ``Proximal {Policy} {Optimization} {Algorithms},'' 2017, arXiv:1707.06347.

\bibitem{ho_generative_2016}
J.~Ho and S.~Ermon, ``Generative {Adversarial} {Imitation} {Learning},'' 2016, arXiv:1606.03476.

\bibitem{paul_post_2020}
M.~Paul and I.~Ghosh, ``Post encroachment time threshold identification for right-turn related crashes at unsignalized intersections on intercity highways under mixed traffic,'' \emph{Int. Journal of Injury Control and Safety Promotion}, vol.~27, no.~2, pp. 121--135, 2020.

\end{thebibliography}
